%% file: main.tex
\documentclass[10pt]{article} 
\usepackage[preprint]{tmlr}

\input{math_commands.tex}

\usepackage{hyperref}

\usepackage{amssymb}
\usepackage{mathtools}
\usepackage{mathrsfs}
\mathtoolsset{showonlyrefs}
\usepackage{graphicx}
\usepackage{subcaption}
\usepackage[space]{grffile}
\usepackage{url}
\usepackage{float}

\title{Plasticity of Growing and Elastic Neural Networks in \\ Online Continual Learning}

\author{\name Jeong Min Kong \email \textnormal{jeongminkong@ucla.edu} \\
      \addr \textnormal{Department of Electrical and Computer Engineering} \\
      \textnormal{University of California, Los Angeles (UCLA)}
      \AND
      \name Richard S. Sutton \email \textnormal{rsutton@ualberta.ca} \\
      \addr \textnormal{Department of Computing Science} \\
      \textnormal{University of Alberta} \\
      \textnormal{Alberta Machine Intelligence Institute (Amii)}}

\def\month{MM}  
\def\year{YYYY} 
\def\openreview{\url{https://openreview.net/forum?id=XXXX}} 

\begin{document}

\maketitle

\begin{abstract}
Neural networks that can grow or both grow and shrink during learning, referred to as \emph{growing neural networks} and \emph{elastic neural networks}, respectively, have recently been explored in offline continual learning with a particular focus on catastrophic forgetting. Driven by the observations that 1) online continual learning closely resembles how animals learn; 2) loss of plasticity --- the progressive decline in a learning network’s ability to learn --- is another crucial challenge facing continual learning; and 3) incremental introduction of randomly initialized hidden units was recently shown to help preserve plasticity, in this paper, we study the plasticity of several foundational growing and elastic networks in online continual learning. Our experiments in supervised learning settings show that adaptive growing networks, which incrementally incorporate new, randomly initialized units to the network while keeping all existing connections adaptive, can maintain high prediction accuracy without losing plasticity \emph{despite} the continuous increase in the dead hidden unit proportion. Furthermore, we demonstrate that adaptive elastic networks, which in addition to progressively adding new hidden units \emph{also prune} estimated dead hidden units at the beginning of each new task, can achieve excellent accuracy without loss of plasticity \emph{while simultaneously} maintaining a near-constant, compact size. Our results suggest that growing and elastic networks, which exhibit the ability to adapt its structure to the relevant learning objectives, can be a promising class of algorithms also for preserving high plasticity in online continual learning.
\end{abstract}

\section{Introduction}

It is standard practice for modern neural networks to have a fixed structure throughout learning, in which the arrangement of neurons/units and their interconnections remain unchanged. Nevertheless, there has been active research on what we call \emph{constructive neural networks} --- learning algorithms that can incrementally remove or add units from the network --- driven by the motivation that the network structure should adapt according to the nature of the learning goal (e.g., complexity) rather than being fixed a priori. Many prior works focused on improving computational efficiency through \emph{pruning}, whereby units deemed not significantly useful (e.g., redundant) are incrementally removed from an initially-\emph{large} network \citep{nnpruning1, nnpruning2, skeleton, sensitivity, datafree, networktrimming, noiseout, apoptosis, annextractknowledge, squeeze}. Other works have explored algorithms that, in contrast, progressively \emph{add} units to an initially-\emph{small} network, which we refer to as \emph{growing neural networks}, or that can both grow and shrink in size, which we call \emph{elastic neural networks}.

Two of the earliest and most notable growing networks are the cascade-correlation learning architecture (CasCor) \citep{cascor} and the neuroevolution of augmenting topologies (NEAT) \citep{neat, neuroevolutionbook}. CasCor is a supervised learning algorithm that begins with a simple linear network and incrementally adds new hidden units, one at a time, until satisfactory performance is reached, whereas NEAT is a genetic algorithm that progressively introduces new units and connections through mutations over successive generations until a target fitness is achieved. These foundational approaches acted as a starting point leading to numerous subsequent algorithms, including direct extensions such as the recurrent cascade-correlation architecture \citep{recurrentcascor}, deep cascade learning \citep{deepcascadelearning}, HyperNEAT \citep{hyperneat}, DeepNEAT and CoDeepNEAT \citep{deepneat}, but also a broader class of growing and elastic networks. More recently, renewed interest in continual learning has motivated the development of gradient-based learning algorithms designed to address lifelong adaptation to new, incoming learning tasks, including dynamically expandable networks \citep{den}, learn-to-grow and learn-to-remember frameworks \citep{learntogrow, learntoremember}, compacting, picking and growing algorithm \citep{cpg}, and progressive neural networks \citep{pnn}. However, these methods primarily focus on 1) \emph{offline} continual learning settings, where, for instance, the full dataset for a supervised learning task is assumed to be available through the entire task-learning duration; and/or 2) mitigating the problem of \emph{catastrophic forgetting}, which is the tendency of a network to lose previously learned knowledge when it is trained on new learning tasks. Motivated by the observations that 1) \emph{online} continual learning, in which data arrive sequentially one sample at a time, closely resembles natural learning systems; 2) \emph{loss of plasticity} --- the gradual degradation of a network’s ability to learn \citep{lop} --- is a challenge in continual learning that is as critical as catastrophic forgetting; and 3) incremental introduction of randomly initialized units has been shown to help preserve plasticity \citep{lop}, in this paper, we focus on studying the plasticity of growing and elastic networks derived from the foundational CasCor algorithm in online continual learning settings.

\section{Experiments}

\subsection{Online Continual Learning Set-Ups}

In this work, we consider two online continual (multi-task) supervised-learning set-ups inspired from \citet{lop}, namely online permuted MNIST and online permuted FashionMNIST (FMNIST), to study the plasticity of various growing and elastic networks. Online permuted MNIST or FMNIST is a sequence of image classification tasks, where each task applies its unique, random pixel permutation to all of the images within the dataset normalized to a value between 0 and 1. The deep learning network is trained in an online fashion; that is, the samples in each task are presented to the network one by one, and after each forward pass, the parameters of the network are immediately optimized to minimize the cross-entropy loss via backpropagation. For all algorithms utilized in this paper, stochastic gradient descent (SGD) with a step size of 0.001 is employed as the optimizer. Note that as pixel permutations are independent across tasks, there is minimal spatial structure shared across tasks; nevertheless, it is possible for a network to learn other, non-spatial features that would facilitate faster or improved learning in subsequent tasks, such as the label structure and statistical patterns of the input activations.

For both online permuted MNIST and FMNIST, we consider two scenarios each with a different number of samples per task. A key motivation for this is to examine how the learning behaviors may vary under different learning durations. Specifically, we consider $N = 10,000$ and $40,000$ samples per task. To build such smaller datasets, we select the first $N/10$ images from the downloaded dataset corresponding to each class, for all classes.

Throughout the paper, two of the most important measurements presented are \emph{accuracy} and \emph{dormancy percentage}. Here, accuracy for a task is defined as the number of correct predictions made during online training divided by the number of samples per task, $N$. We define dormancy percentage for a task as the number of constant or ``dead'' hidden units in the learning network divided by the total number of hidden units in the network at the beginning of the task. As described in \citet{lop}, a unit is considered constant if the weights coming into the unit never change or very slightly change due to only zero or near-zero gradients. When a unit employs ReLU as the activation function, which is the case for all algorithms utilized in this paper, this happens when a unit only outputs 0 for all samples of the task, or equivalently, is ``dead''. When computing the dormancy percentage, we \emph{approximate} whether a hidden unit is dead using only a tiny random subset of the dataset, i.e., 5\% of $N$. We note that this metric is particularly interesting to monitor in the context of our work, as the experiments in \citet{lop} observed a correlation between declining accuracy and loss of plasticity across tasks with increasing dormancy percentage.

In the remainder of this paper, we examine the plasticity of various deep learning algorithms on the described online permuted MNIST and FMNIST set-ups. We begin with the standard, fixed fully-connected neural networks as a baseline, followed by a set of growing and elastic networks, namely staged growing networks, adaptive growing networks, adaptive elastic networks, and two-layer variants of adaptive growing and elastic networks. The experimental results presented in the main section of this paper are for online permuted MNIST and $N = 10,000$ samples per task; results for other experiments are provided in the appendix.

\subsection{Fixed Fully-Connected Neural Networks (F-FCNNs)}

\begin{figure}[!t]
    \centering
    \vspace{0.0in}    \includegraphics[width=0.8\linewidth,trim=0 15mm 0 15mm,clip]{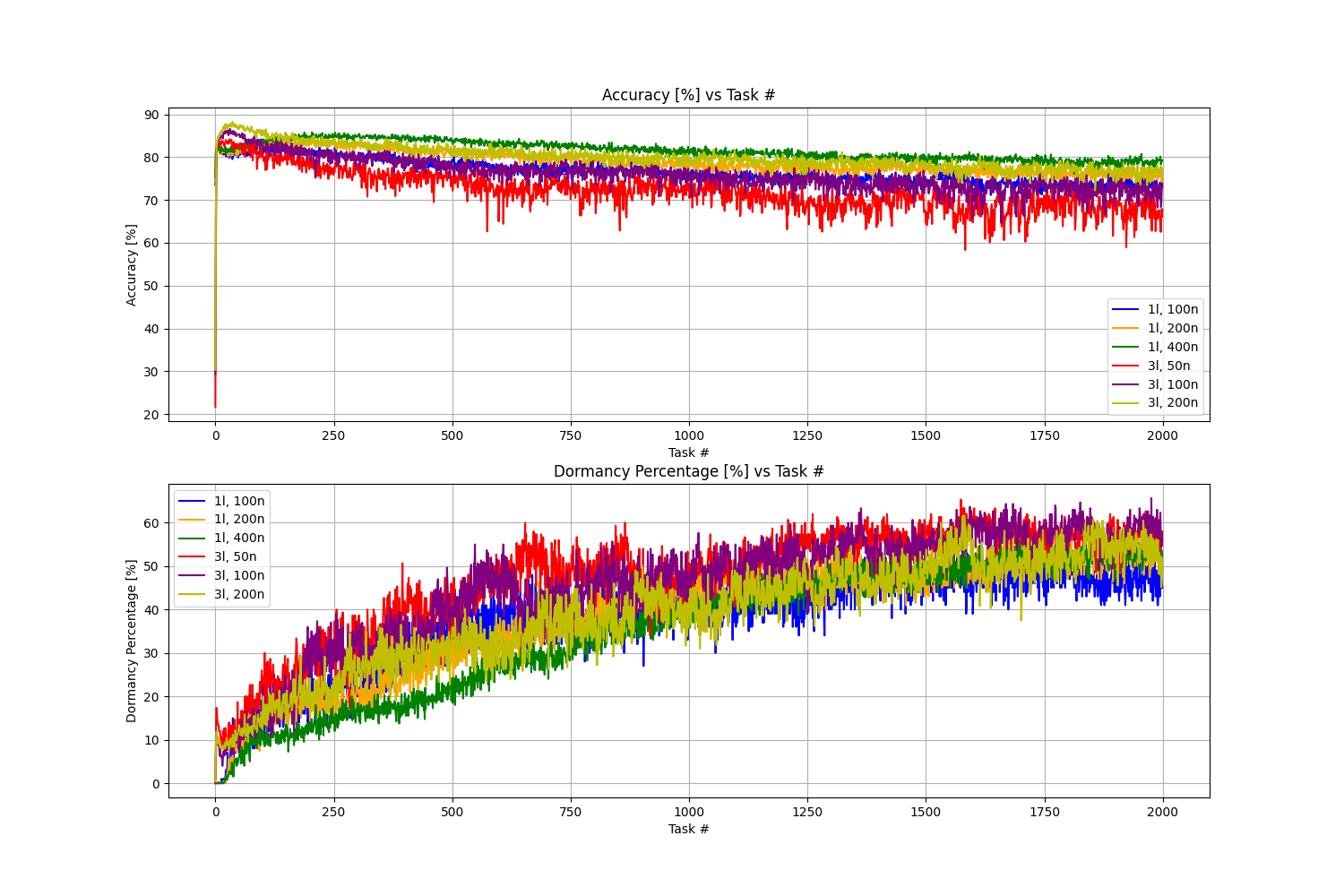}
    \caption{Accuracy and dormancy \% plots for F-FCNNs with varying number of hidden layers ($l$) and hidden sizes ($n$).}
    \label{fig:FFCNN_acc}
\end{figure}

As the name suggests, fixed fully-connected neural networks (F-FCNNs) are dense neural networks that, unlike constructive networks, maintain a fixed network structure throughout the online continual learning process. Figure~\ref{fig:FFCNN_acc} presents the accuracy and dormancy percentage plots for F-FCNNs with varying depths and hidden sizes. As demonstrated in \citet{lop}, the models progressively lose plasticity, reflected by the steady decline in accuracy over successive tasks. It can also be seen that the dormancy percentage increases or saturates at a high value.

\subsection{Staged Growing Networks (SGNs)}

\begin{figure}[!t]
    \centering
    \vspace{0.0in}    \includegraphics[width=0.8\linewidth,height=\textheight,keepaspectratio]{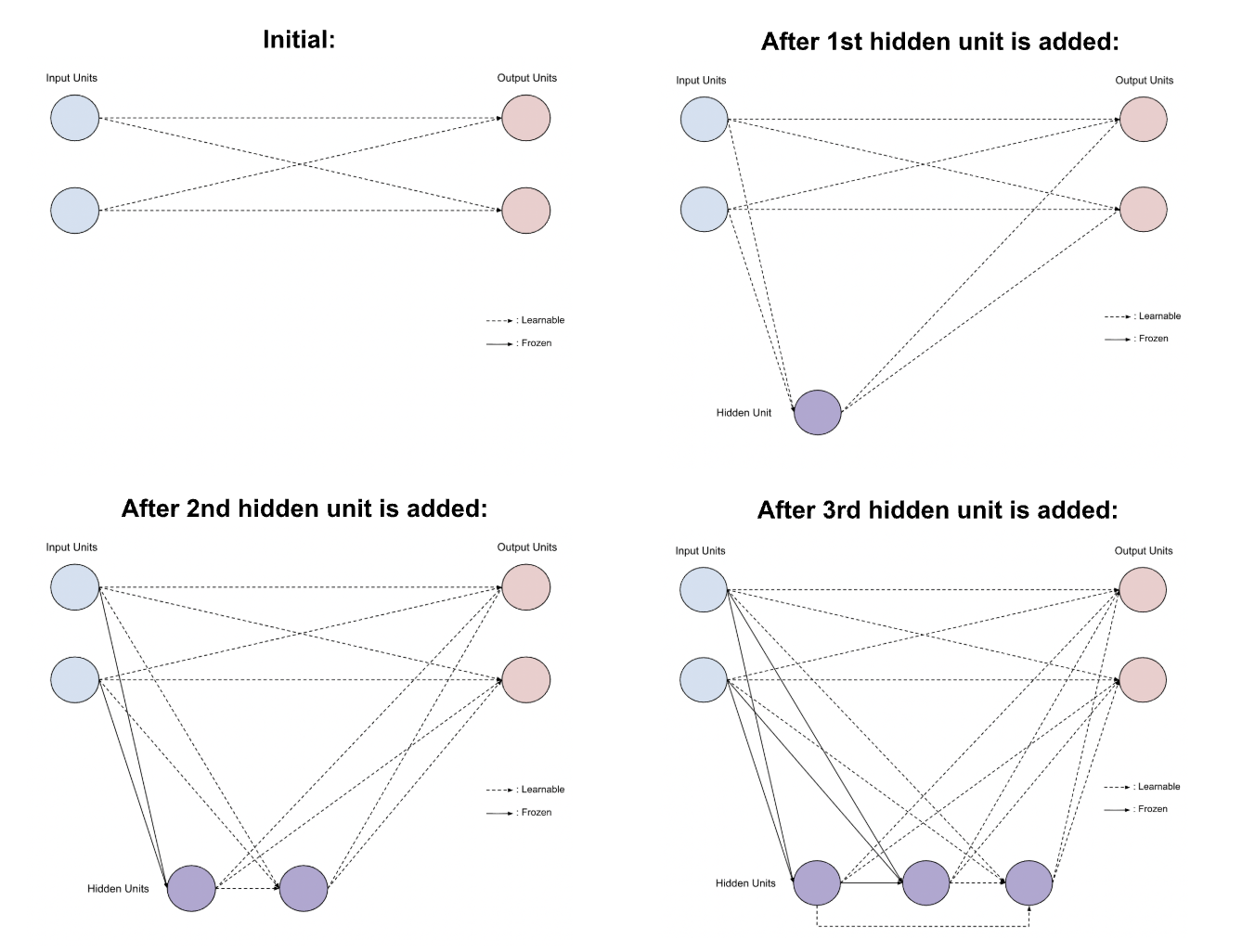}
    \caption{Visualization of SGN as new hidden units are added to the network. Observe that the connections between the input and output units and the outgoing connections from the hidden units to the output units are always adaptive.}
    \label{fig:SCN_Plot}
\end{figure}

\begin{figure}[!t]
    \centering
    \vspace{0.0in}    \includegraphics[width=0.8\linewidth,trim=0 15mm 0 15mm,clip]{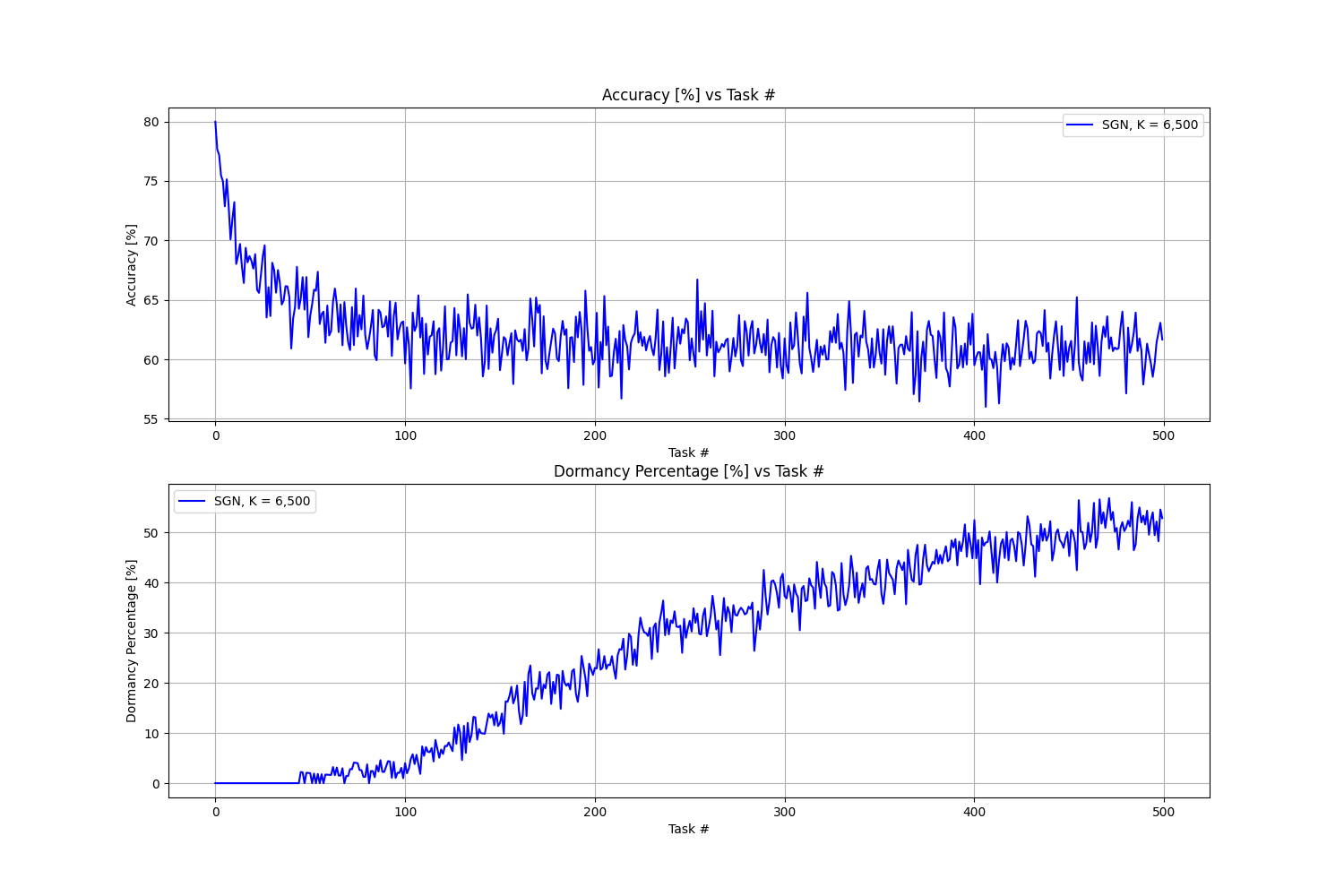}
    \caption{Accuracy and dormancy \% plots for SGN with $K = 6,500$ (one new hidden unit per task).}
    \label{fig:SCN_acc}
\end{figure}

Staged growing network (SGN), which we illustrate in Figure~\ref{fig:SCN_Plot}, is a variant of the CasCor algorithm. In SGN, online continual learning begins with a simple linear network that densely connects the input units to the output units. The network parameters are optimized to minimize the prediction error using backpropagation. After the first $K$ samples in the task, learning is temporarily halted, and a new hidden unit is incorporated into the network. Once the input units are connected to the input of this new hidden unit and its output is connected to the output units, with randomized weights (e.g., Kaiming initialization or uniform random initialization), the learning resumes. After another $K$ samples, again, the learning is paused, and a new hidden unit is added. The new hidden unit forms connections with the input and output units as before, but furthermore, the incoming connections of the existing, previously-added hidden unit are frozen, and its output is additionally connected to the new hidden unit, with a randomized weight. The above procedure repeats every $K$ samples until the end of the task, with the parameters that were once frozen remaining frozen throughout the learning process, and \emph{all} of the existing hidden units always forming connections with the new hidden unit. As the features are learned sequentially, one at a time, we call this algorithm ``staged'' growing networks. Once the current task is completed, the model is carried over to the next task, where learning proceeds from its latest state and a new hidden unit is introduced every $K$ samples. We highlight that the main differences between SGN and CasCor are 1) CasCor is designed for batch learning, whereas SGN is also suitable for online learning; and 2) CasCor learns features by maximizing the correlation between the candidate hidden units and the residual error, while SGN learns features by minimizing the prediction error using backpropagation. SGN is similar to the constructive network introduced in \citet{javed_paper} and \citet{javed_thesis}, the notable differences being that in their architecture the hidden units have recurrent connections, and there are no direct connections between the input units and the output units.

Figure~\ref{fig:SCN_acc} presents the accuracy and dormancy percentage plots for SGN with $K = 6,500$ samples (i.e., one new hidden unit per task). It can be seen that the SGN loses plasticity, as evidenced by the declined accuracy. This behavior is unsurprising, as the model is forced to not only utilize fixed features learned from previous tasks that remain useful, but \emph{also} features that are no longer relevant to the current task. Consequently, as new hidden units are introduced across tasks and the accumulated noisy features increase, a reduction in plasticity is expected. We speculate that the decline eventually plateaus once the network capacity becomes sufficiently large to offset the increasing noisy features. Another interesting observation is the continuous rise in dormancy percentage, which indicates that a newly added hidden unit often just dies before it becomes frozen. The notable trends from Figure~\ref{fig:SCN_acc} can be seen in other experiments as well, i.e., online permuted MNIST with $N$ = 40,000 samples per task and online permuted FMNIST with $N$ = 10,000 and 40,000, shown in the appendix.

\subsection{Adaptive Growing Networks (AGNs)}

\begin{figure}[!t]
    \centering
    \vspace{0.0in}    \includegraphics[width=0.8\linewidth,trim=0 15mm 0 15mm,clip]{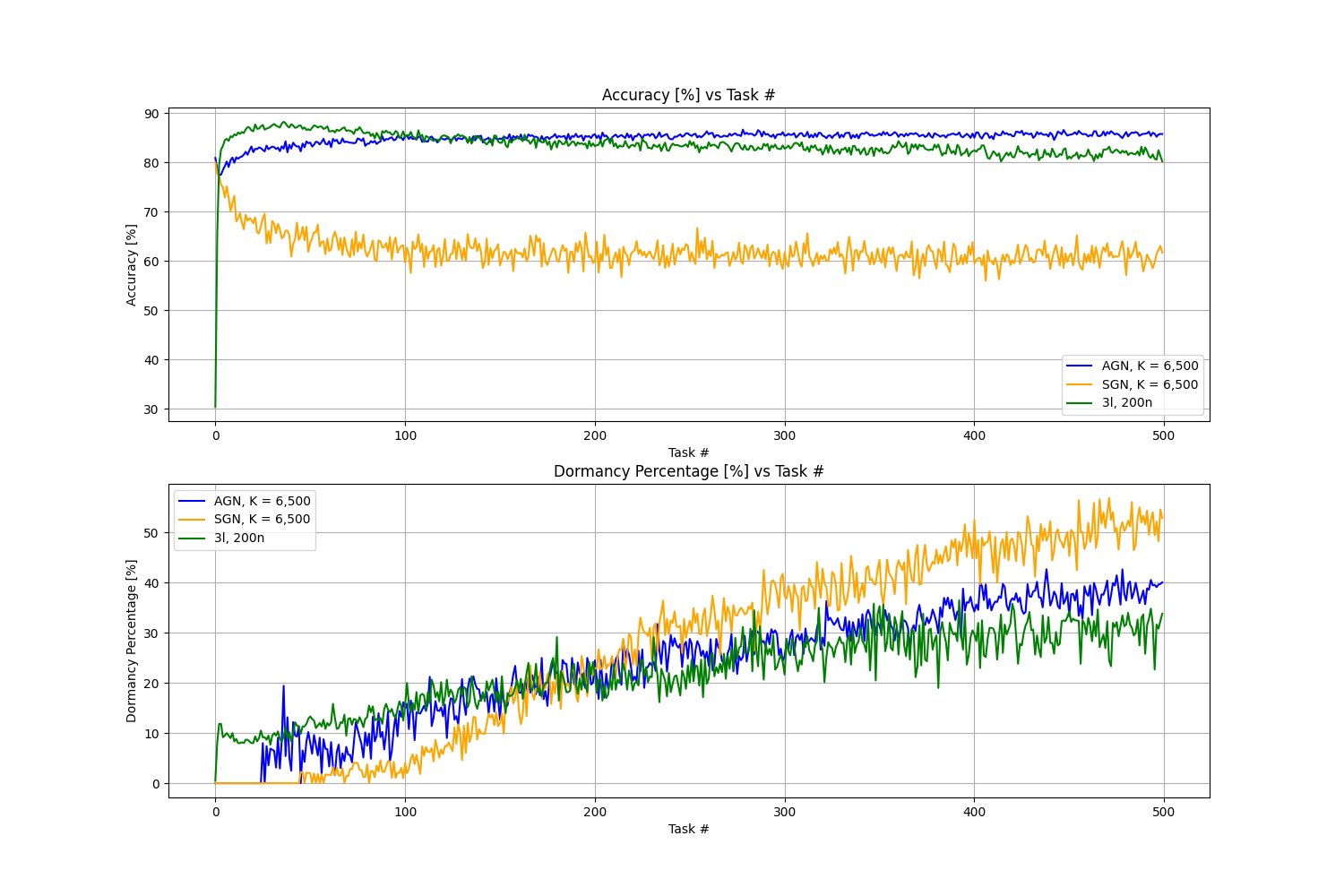}
    \caption{Accuracy and dormancy \% plots for AGN and SGN with $K = 6,500$ (one new hidden unit per task), and F-FCNN with 3 hidden layers and a hidden size of 200.}
    \label{fig:ACN_acc}
\end{figure}

To address a key limitation of SGNs in continual learning settings, namely the enforced use of old features across new tasks, we introduce a simple variation of SGNs called adaptive growing networks (AGNs). AGN is identical to SGN, with the exception that all network connections remain adaptive and are never frozen.

Figure~\ref{fig:ACN_acc} shows the accuracy and dormancy percentage plots for AGN with $K = 6,500$ (i.e., one new hidden unit per task) along with several baselines. From the accuracy plot, it can be seen that the AGN consistently achieves excellent classification performance across tasks without losing plasticity, unlike F-FCNN, which gradually loses plasticity after an early peak. This demonstrates that a simple modification to SGN, namely keeping all connections always adaptive, can enable the network to maintain high plasticity and accuracy. A further interesting observation can be made from the dormancy percentage plot. Despite AGNs maintaining high accuracy throughout, the dormancy percentage \emph{continually increases}. This differs from earlier algorithms, where increasing dormancy percentage was coupled with declining performance. We hypothesize that this distinct behavior results from the algorithm continuously adapting some of the existing connections to new tasks while progressively injecting new learning capacity through randomly initialized units, in a manner similar to continual backpropagation \citep{lop}. The notable trends in Figure~\ref{fig:ACN_acc} are also apparent in other experiments.

One major drawback of AGN is that the number of hidden units is always growing, as new hidden units are repeatedly incorporated into the network without ever being removed. This not only rapidly increases memory usage due to the exponentially growing number of connections, but also the processing time of both forward and backward passes, since adding a new hidden unit is equivalent to adding a new single-unit hidden layer. These limitations motivate the development of an AGN variant that can continuously prune units and sustain a compact size while still being able to maintain high plasticity and accuracy.

\subsection{Adaptive Elastic Networks (AENs)}

\begin{figure}[!t]
    \centering
    \vspace{0.0in}    \includegraphics[width=0.9\linewidth,height=\textheight,keepaspectratio]{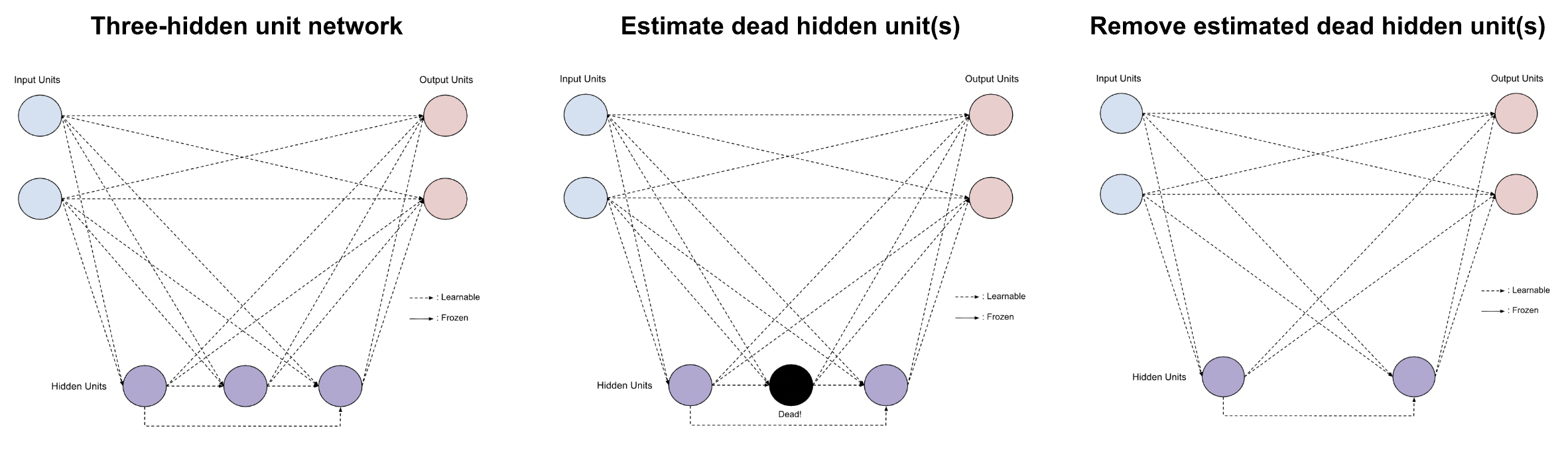}
    \caption{Visualization of the pruning process in AEN.}
    \label{fig:SACN_Plot}
\end{figure}

\begin{figure}[!t]
    \centering
    \vspace{0.0in}    \includegraphics[width=0.8\linewidth,trim=0 15mm 0 15mm,clip]{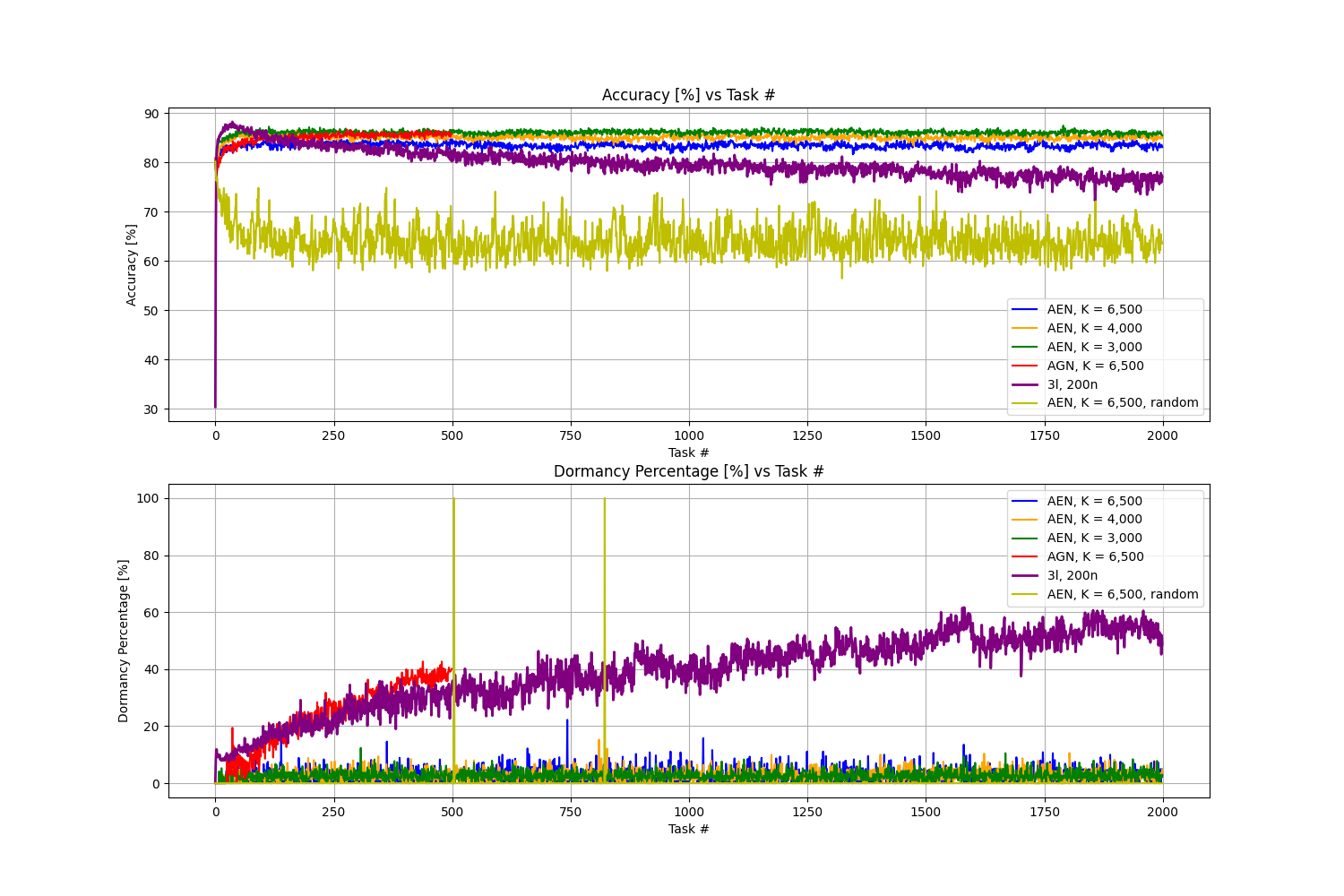}
    \caption{Accuracy and dormancy \% plots for AENs with $K = 6,500$ (one new hidden unit per task), $K = 4,000$ (two new hidden units per task), and $K = 3,000$ (three new hidden units per task), along with several baselines. AEN, $K = 6,500$, random, like AEN, $K = 6,500$, adds one new hidden unit per task, but it employs a different pruning strategy. Specifically, at the beginning of each task, 1) the number of hidden units to remove is randomly selected (0 included); and 2) that number of hidden units is then randomly chosen and removed.}
    \label{fig:SACN_acc}
\end{figure}

\begin{figure}[!t]
    \centering
    \vspace{0.0in}    \includegraphics[width=0.8\linewidth,trim=0 15mm 0 15mm,clip]{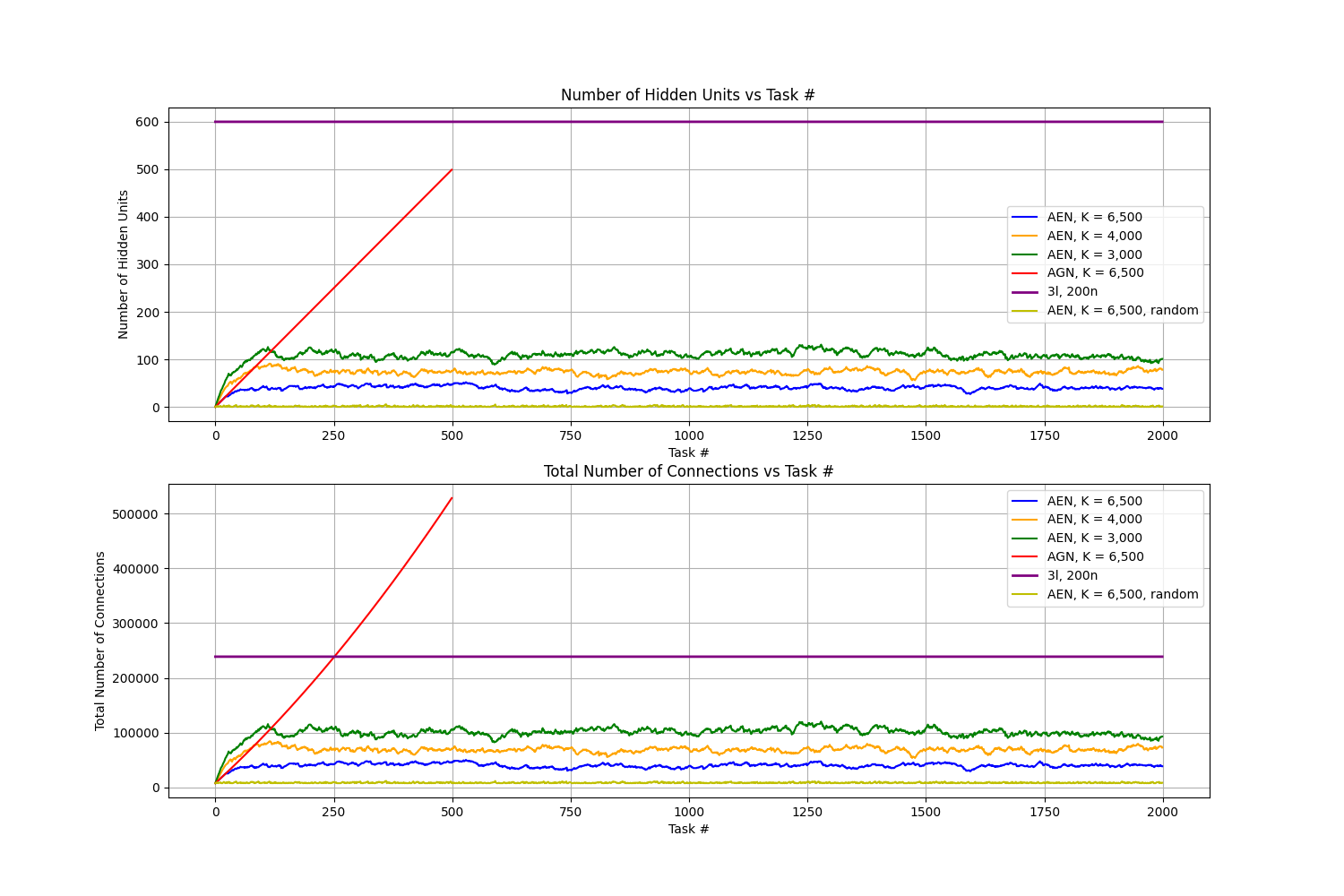}
    \caption{Number of hidden units and number of connections plots for AENs with $K = 6,500$ (one new hidden unit per task), $K = 4,000$ (two new hidden units per task), and $K = 3,000$ (three new hidden units per task), along with several baselines.}
    \label{fig:SACN_size}
\end{figure}

Adaptive elastic network (AEN) is an extension of AGN that introduces periodic pruning of hidden units. Specifically, at the beginning of each new task, hidden units that are \emph{estimated} to be dead are removed. Note that if only the dead hidden units are removed, AENs would have the same learning trajectory as the AGNs. However, because \emph{estimated} dead hidden units are removed here, some active hidden units would be pruned as well; this leads to a different learning trajectory than the AGNs. Figure~\ref{fig:SACN_acc} presents the accuracy and dormancy percentage plots for AENs with $K = 6,500$ (i.e., one new hidden unit per task), $K = 4,000$ (i.e., two new hidden units per task), and $K = 3,000$ (i.e., three new hidden units per task). Consistent with the dormancy percentage calculation, the dead hidden units to be pruned are estimated using a random 5\% of the samples for each new task.

It can be observed that all three AENs maintain high accuracy without losing plasticity, with performance improving as more hidden units are added per task. Additionally, the dormancy percentage of AENs consistently hovers around 0\%, in contrast to that of AGN, which always increases. This consequently implies that unlike AGNs, AENs utilize the network parameters with high efficiency. Perhaps, the most interesting question arising from employing a pruning strategy is: how does the network size evolve over time? Figure~\ref{fig:SACN_size} presents the number of hidden units and total number of connections across tasks for AENs. Fascinatingly, it can be observed that the network \emph{converges} to a compact size beyond a certain point, with the saturation level increasing with more new hidden units per task. This intriguing finding, together with earlier observations from Figure~\ref{fig:SACN_acc}, suggests that a simple pruning strategy, of removing estimated dead hidden units at the beginning of each new task, can enable the network to retain high plasticity and accuracy \emph{while also} consistently maintaining a compact size.

Notable trends from Figure~\ref{fig:SACN_acc} and Figure~\ref{fig:SACN_size} are also evident in other experiments, though some interesting patterns can be observed across them. First, the saturated size appears to scale approximately linearly with the number of new units per task for online permuted MNIST, i.e., doubling the number of new units per task roughly doubles the saturated size, etc, while not for online permuted FMNIST. Second, for the case of one new unit per task, the difference in the saturated size between $N = 10,000$ and $N = 40,000$ appears to be substantial for online permuted FMNIST, while minimal for online permuted MNIST. Moreover, even though FMNIST is a more difficult classification task than MNIST, there are several cases when the saturated size for online permuted FMNIST is lower than the online permuted MNIST counterpart.

\subsection{Two-Layer Growing and Elastic Networks}

\begin{figure}[!t]
    \centering
    \vspace{0.0in}    \includegraphics[width=0.85\linewidth,height=\textheight,keepaspectratio]{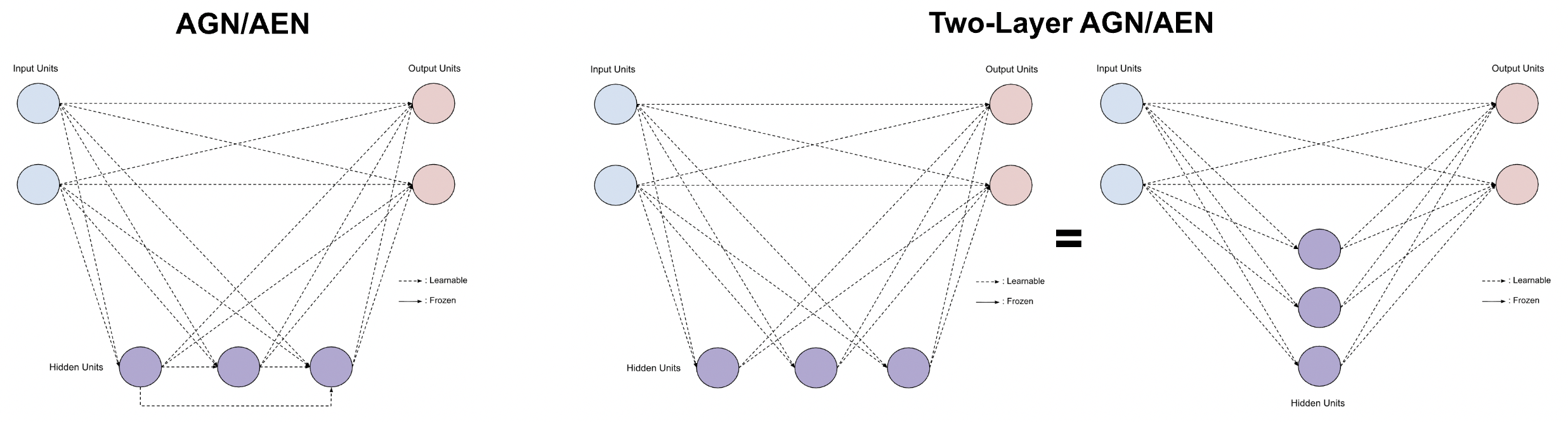}
    \caption{Visualization of AGN/AEN and their two-layer variant.}
    \label{fig:Two-layer_Plot}
\end{figure}

\begin{figure}[!t]
    \centering
    \vspace{0.0in}    \includegraphics[width=0.77\linewidth,trim=0 35mm 0 35mm,clip]{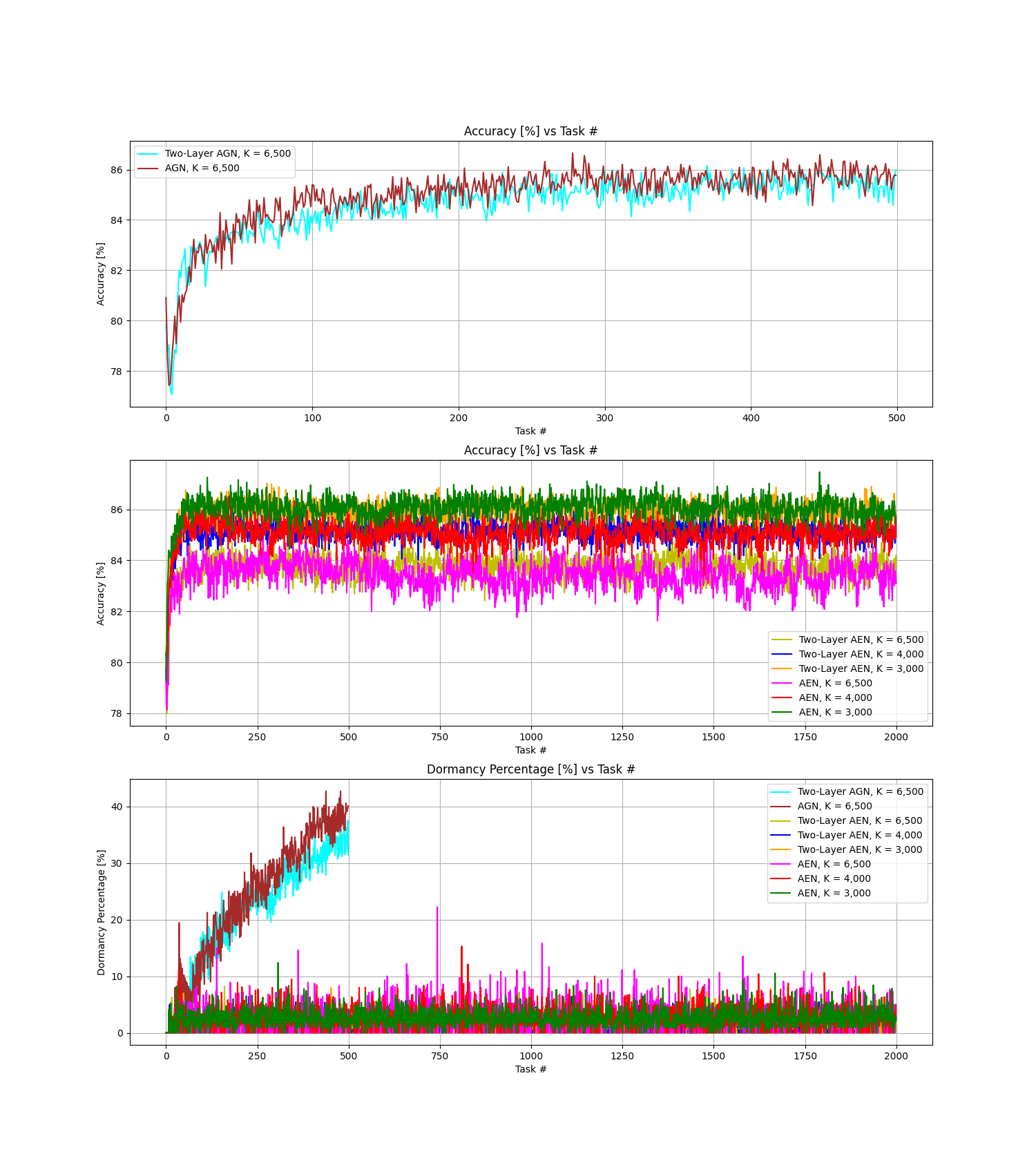}
    \caption{Accuracy and dormancy \% plots for two-layer AGN with $K = 6,500$ (one new hidden unit per task), two-layer AENs with $K = 6,500$, $K = 4,000$ (two new hidden units per task), and $K = 3,000$ (three new hidden units per task), along with their original counterparts.}
    \label{fig:two-layer_acc}
\end{figure}

\begin{figure}[!t]
    \centering
    \vspace{0.0in}    \includegraphics[width=0.8\linewidth,trim=0 15mm 0 15mm,clip]{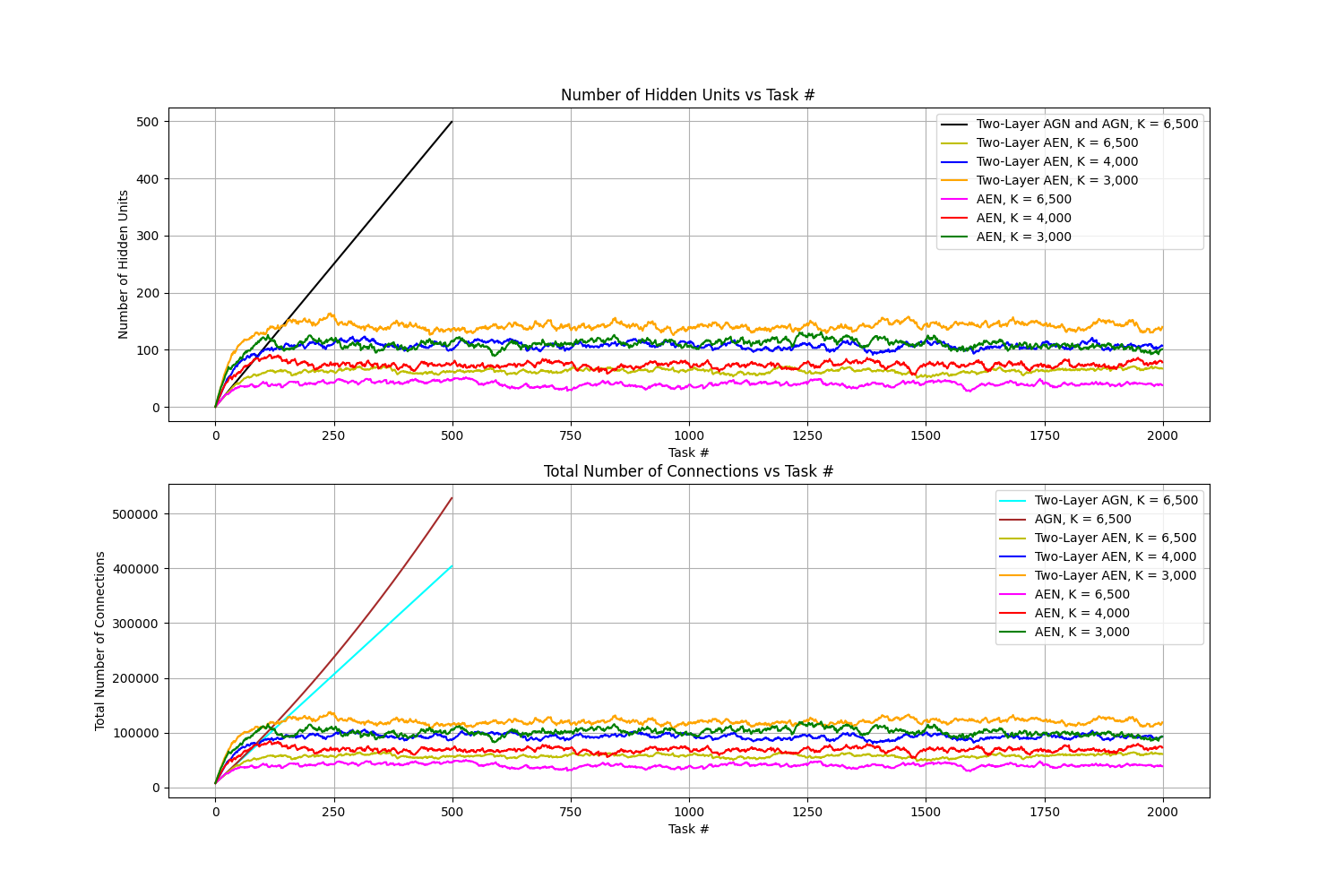}
    \caption{Number of hidden units and number of connections plots for two-layer AGN with $K = 6,500$ (one new hidden unit per task), two-layer AENs with $K = 6,500$, $K = 4,000$ (two new hidden units per task), and $K = 3,000$ (three new hidden units per task), along with their original counterparts.}
    \label{fig:two-layer_size}
\end{figure}

Motivated by the observation that prohibiting any connection between hidden units in AGN and AEN effectively constrains the networks to a two-layer structure, which could substantially reduce the computational time of both the forward and backward passes, we further explore variants called two-layer AGN and two-layer AEN. More specifically, two-layer AGN and two-layer AEN are identical to AGN and AEN, respectively, except that each newly added hidden unit receives connections \emph{only} from the input units and never from existing hidden units.

Figure~\ref{fig:two-layer_acc} presents the accuracy and dormancy percentage plots, and Figure~\ref{fig:two-layer_size} the network size plots, for a two-layer AGN with $K = 6,500$ (i.e., one new hidden unit per task) and two-layer AENs with $K = 6,500$ (i.e., one new hidden unit per task), $K = 4,000$ (i.e., two new hidden units per task), and $K = 3,000$ (i.e., three new hidden units per task). It can be observed that the two-layer variants exhibit similar notable behaviors as their original counterparts, which are also apparent in other experiments. It is important to highlight, however, that the converged network sizes can be larger for the two-layer case than their respective counterparts. For instance, it can be seen that the converged total number of connections is larger for the two-layer case than their original counterparts for online permuted MNIST, while they are approximately the same for online permuted FMNIST. Despite the potential increase, the possibility for more efficient forward and backward passes can still make the two-layer variants a compelling alternative.

\section{Conclusion and Future Works}

This paper investigated the plasticity of several foundational growing and elastic neural networks in online continual supervised-learning settings. Among several key findings, we showed that adaptive growing networks can sustain high predication accuracy without loss of plasticity, despite a continual rise in the fraction of dead hidden units. We further showed that adaptive elastic networks can also consistently achieve high plasticity and accuracy, while simultaneously maintaining a compact, near-constant network size. Overall, our experimental results highlight growing and elastic networks as viable methods for keeping high plasticity in online continual learning.

Future works can be broadly divided into two directions. The first direction involves further exploring the growing and elastic networks presented in this paper, including studying how the timing of adding and pruning units may affect the algorithm’s performance and the network's structural evolution, as well as examining the behaviors of other measures known to be associated with loss of plasticity, such as the effective rank of the representation and average weight magnitude of the network \citep{lop}. The second direction involves the design and analysis of new algorithms, including networks that can incorporate more than one new unit at a time, similar to the constructive-columnar networks from \citet{javed_paper} and \citet{javed_thesis}, and networks that can better exploit both the depth and the hidden size, a gap between the adaptive networks and their two-layer variants from this work.

\section*{Acknowledgments}

Jeong Min Kong is supported by the Amazon AI PhD Fellowship. Jeong Min Kong would like to thank Amazon for generously providing Amazon Web Services (AWS) credits used for this work, and the members of the Reinforcement Learning and Artificial Intelligence (RLAI) lab and the Alberta Plan for AI Research Step 2 meeting at the University of Alberta (especially Edan Meyer, Parash Rahman, Andrew Freeman, Henry Du, and Farzane Aminmansour) for many fruitful discussions.

\bibliography{main}
\bibliographystyle{tmlr}

\appendix

\clearpage
\section{Experimental Results for Online Permuted FMNIST, $N = 10,000$}
\label{sec:appendixA}

\begin{figure}[H]
    \centering
    \vspace{0.0in}    \includegraphics[width=0.95\linewidth,trim=0 15mm 0 15mm,clip]{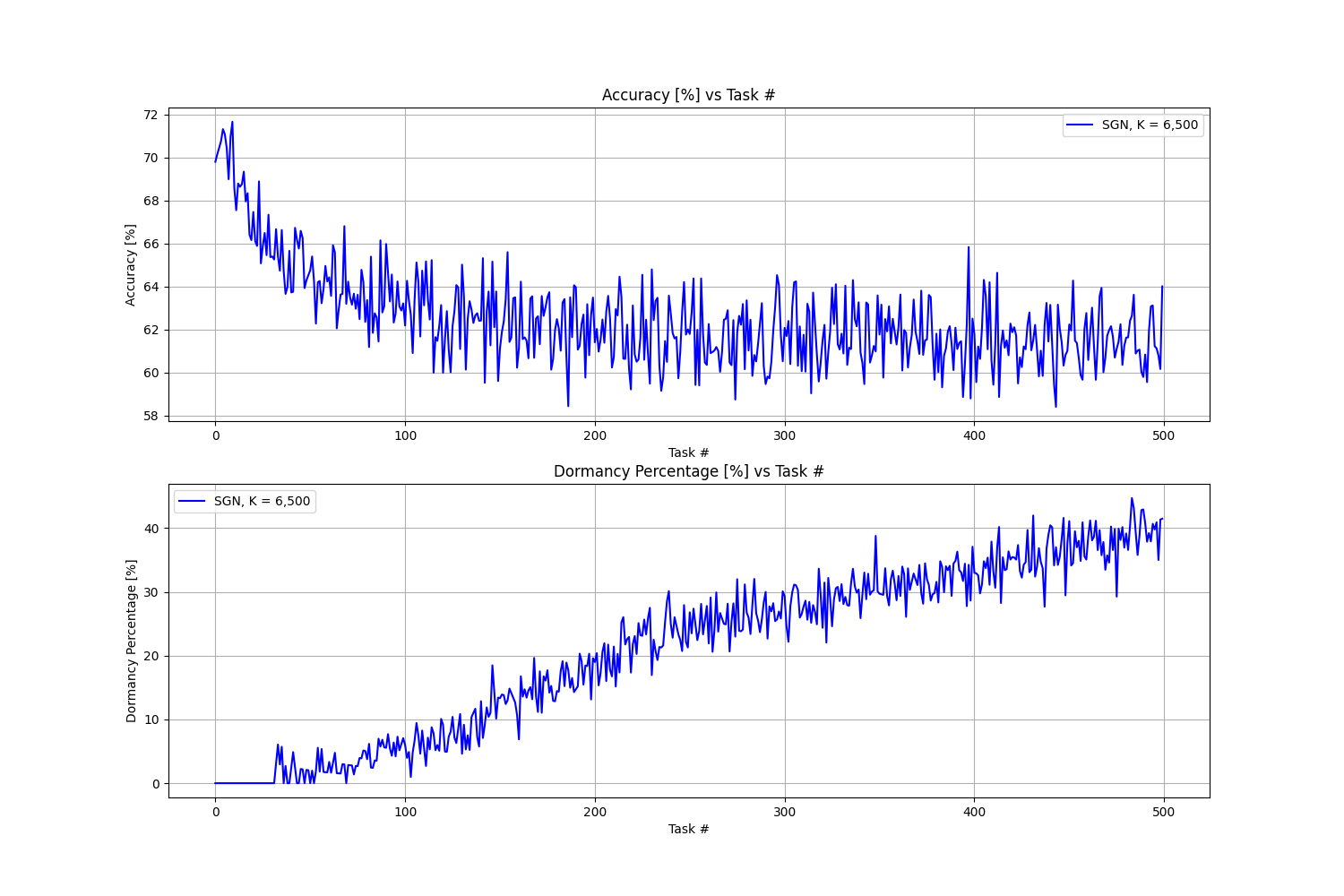}
    \caption{Accuracy and dormancy \% plots for SGN with $K = 6,500$ (one new hidden unit per task).}
    \label{fig:SCN_acc_fmnist10k}
\end{figure}

\begin{figure}[H]
    \centering
    \vspace{0.0in}    \includegraphics[width=0.95\linewidth,trim=0 15mm 0 15mm,clip]{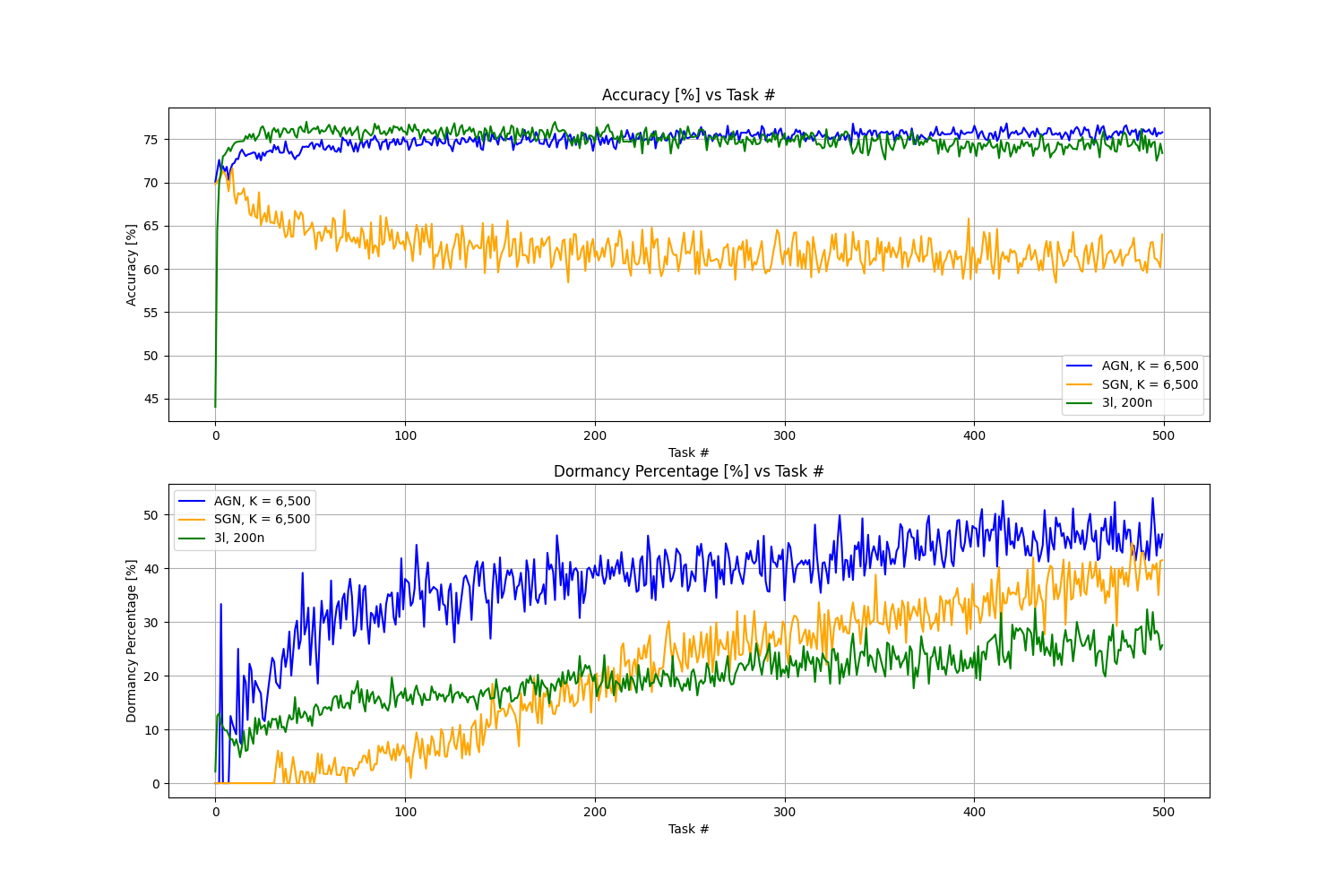}
    \caption{Accuracy and dormancy \% plots for AGN and SGN with $K = 6,500$ (one new hidden unit per task), and F-FCNN with 3 hidden layers and a hidden size of 200.}
    \label{fig:ACN_acc_fmnist10k}
\end{figure}

\begin{figure}[H]
    \centering
    \vspace{0.0in}    \includegraphics[width=0.95\linewidth,trim=0 15mm 0 15mm,clip]{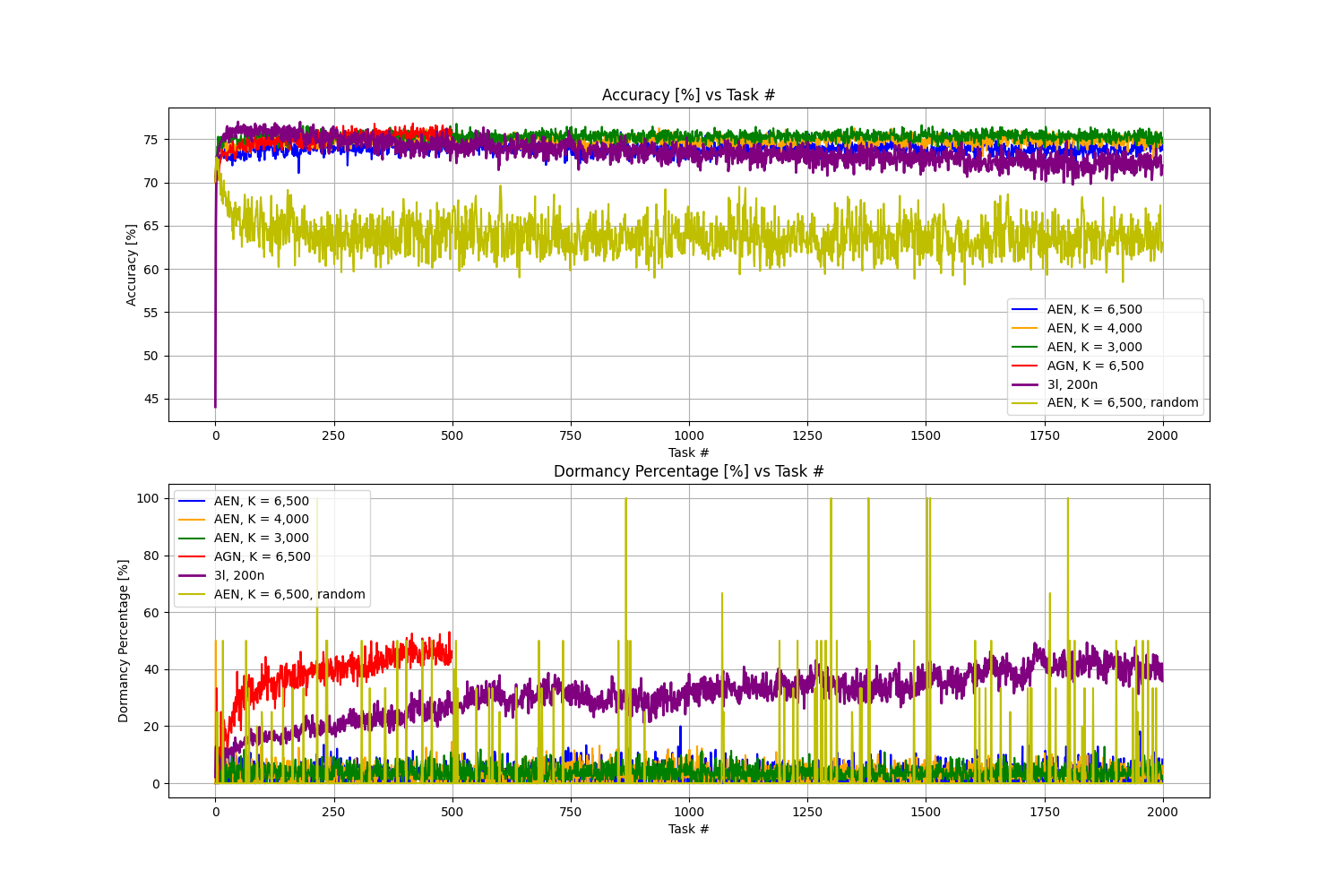}
    \caption{Accuracy and dormancy \% plots for AENs with $K = 6,500$ (one new hidden unit per task), $K = 4,000$ (two new hidden units per task), and $K = 3,000$ (three new hidden units per task), along with several baselines.}
    \label{fig:SACN_acc_fmnist10k}
\end{figure}

\begin{figure}[H]
    \centering
    \vspace{0.0in}    \includegraphics[width=0.95\linewidth,trim=0 15mm 0 15mm,clip]{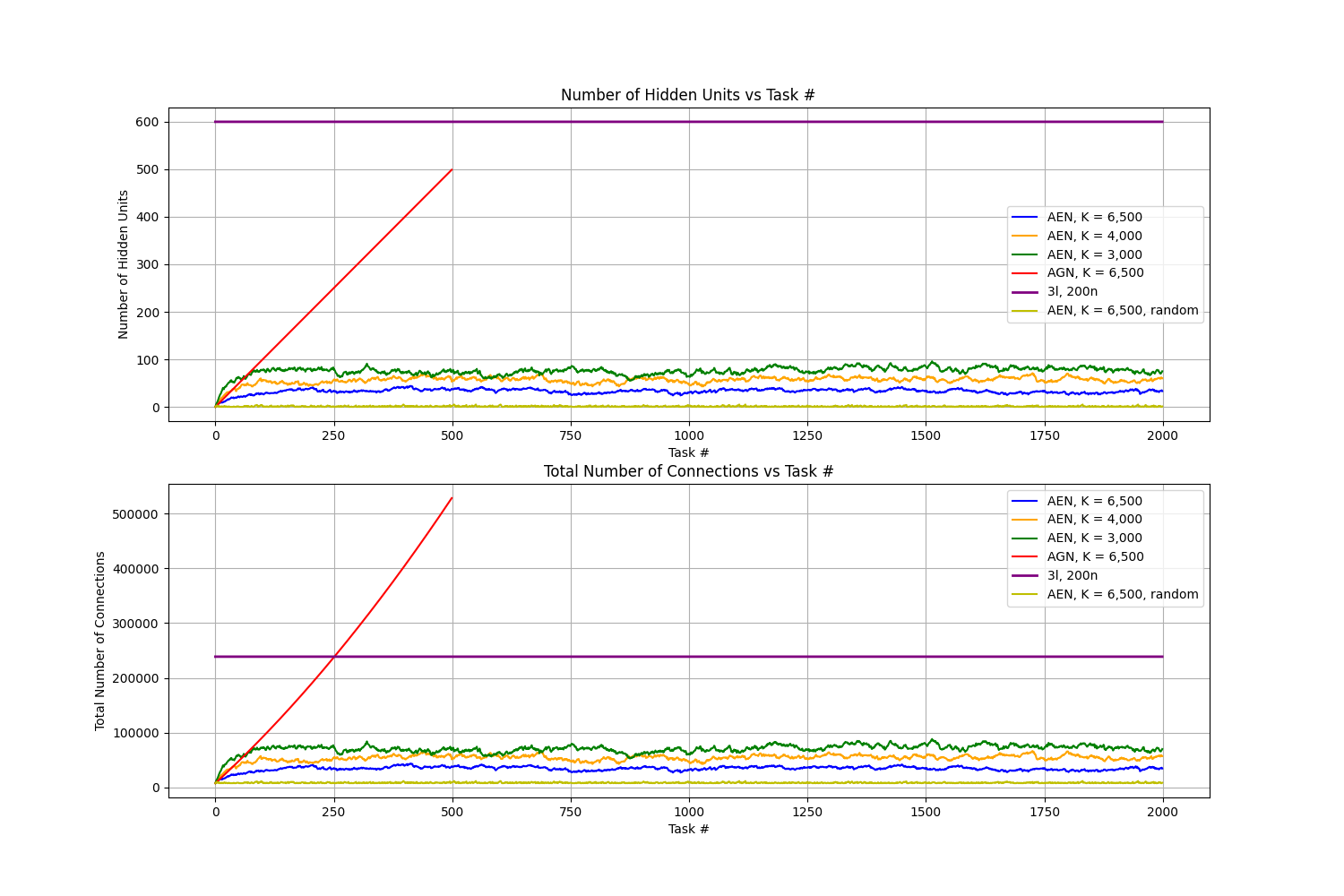}
    \caption{Number of hidden units and number of connections plots for AENs with $K = 6,500$ (one new hidden unit per task), $K = 4,000$ (two new hidden units per task), and $K = 3,000$ (three new hidden units per task), along with several baselines.}
    \label{fig:SACN_size_fmnist10k}
\end{figure}

\begin{figure}[H]
    \centering
    \vspace{0.0in}    \includegraphics[width=0.73\linewidth,trim=0 35mm 0 35mm,clip]{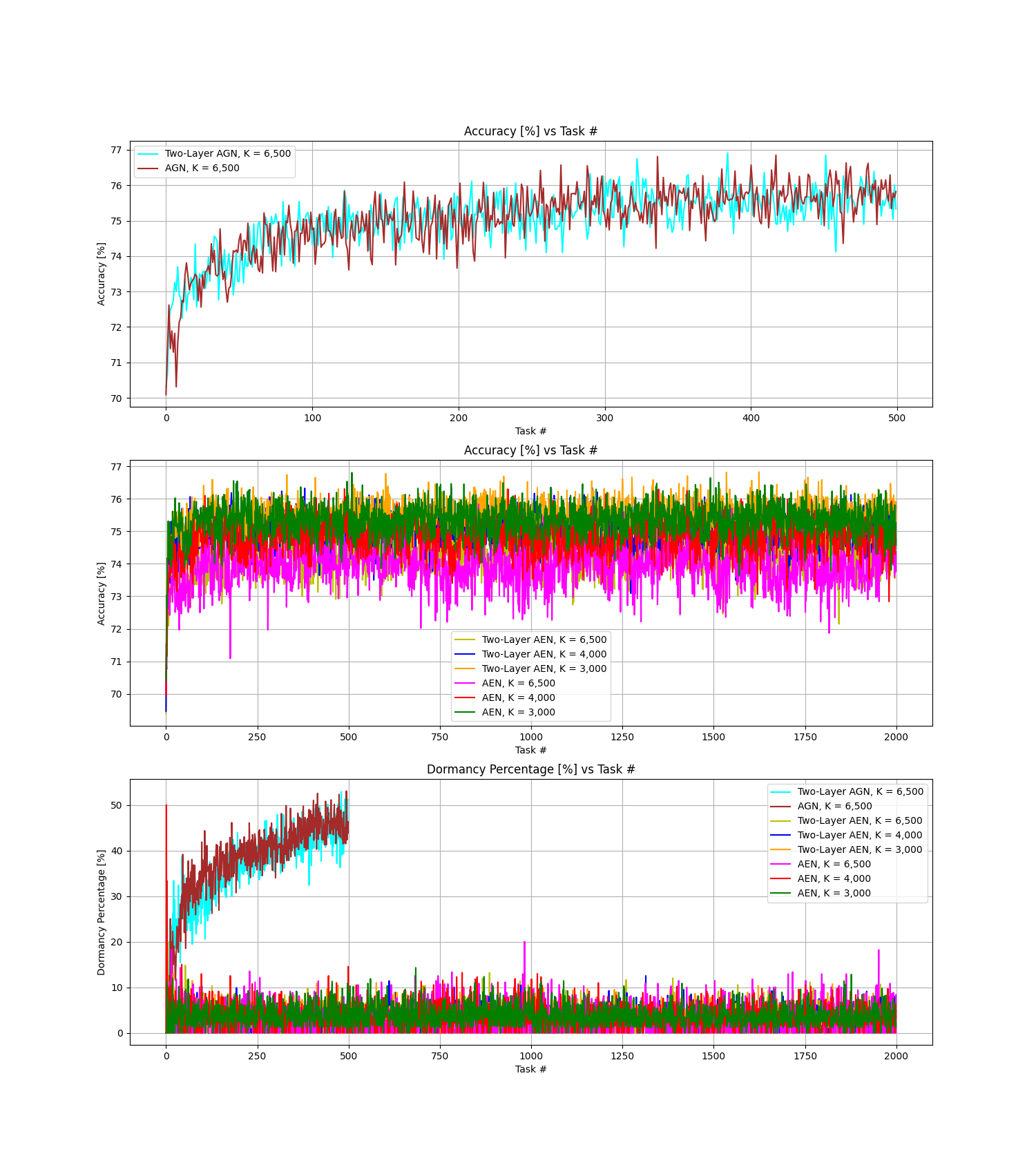}
    \caption{Accuracy and dormancy \% plots for two-layer AGN with $K = 6,500$ (one new hidden unit per task), two-layer AENs with $K = 6,500$, $K = 4,000$ (two new hidden units per task), and $K = 3,000$ (three new hidden units per task), along with their original counterparts.}
    \label{fig:two-layer_acc_fmnist10k}
\end{figure}

\begin{figure}[H]
    \centering
    \vspace{0.0in}    \includegraphics[width=0.73\linewidth,trim=0 15mm 0 15mm,clip]{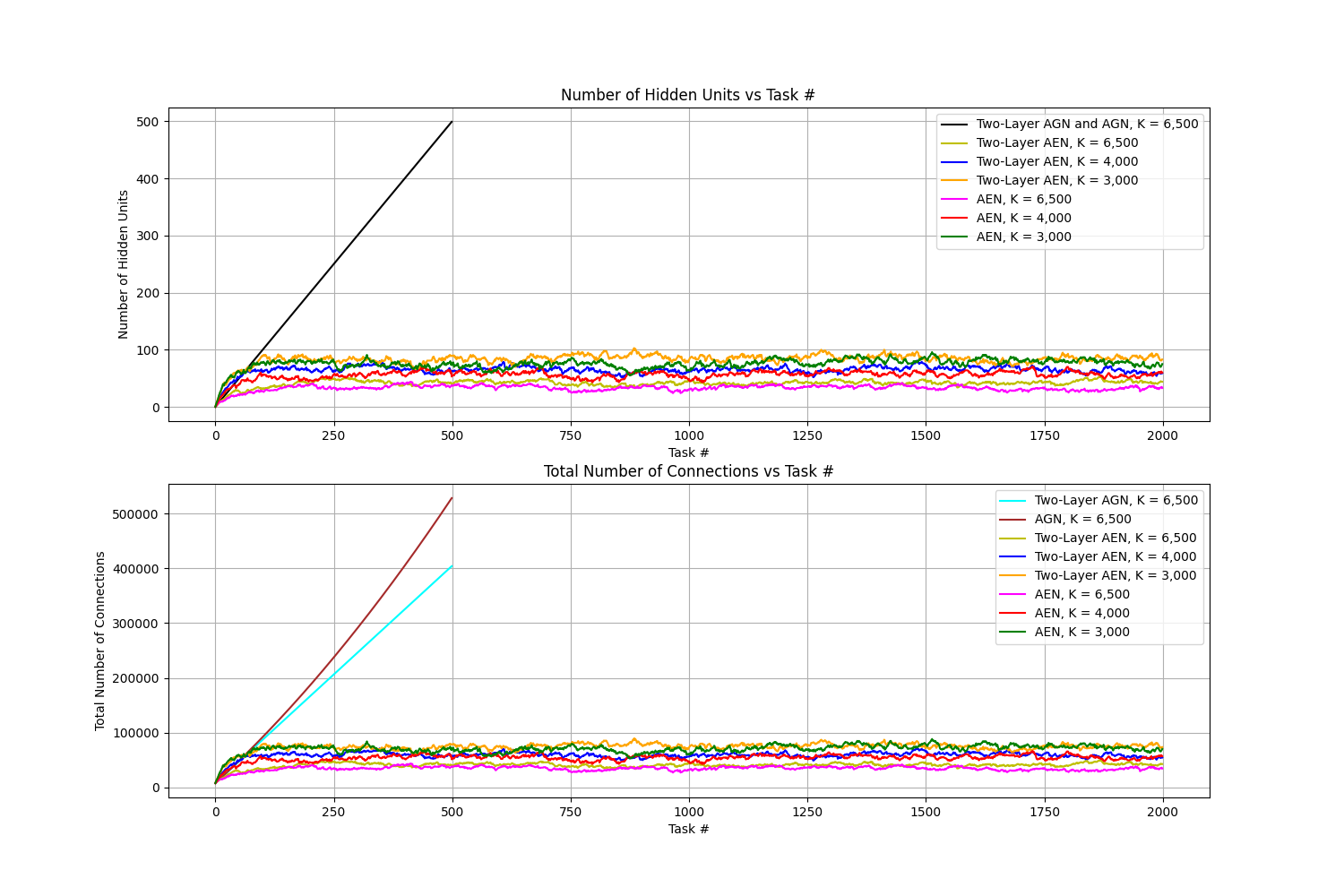}
    \caption{Number of hidden units and number of connections plots for two-layer AGN with $K = 6,500$ (one new hidden unit per task), two-layer AENs with $K = 6,500$, $K = 4,000$ (two new hidden units per task), and $K = 3,000$ (three new hidden units per task), along with their original counterparts.}
    \label{fig:two-layer_size_fmnist10k}
\end{figure}

\clearpage
\section{Experimental Results for Online Permuted MNIST, $N = 40,000$}
\label{sec:appendixB}

\begin{figure}[H]
    \centering
    \vspace{0.0in}    \includegraphics[width=0.95\linewidth,trim=0 15mm 0 15mm,clip]{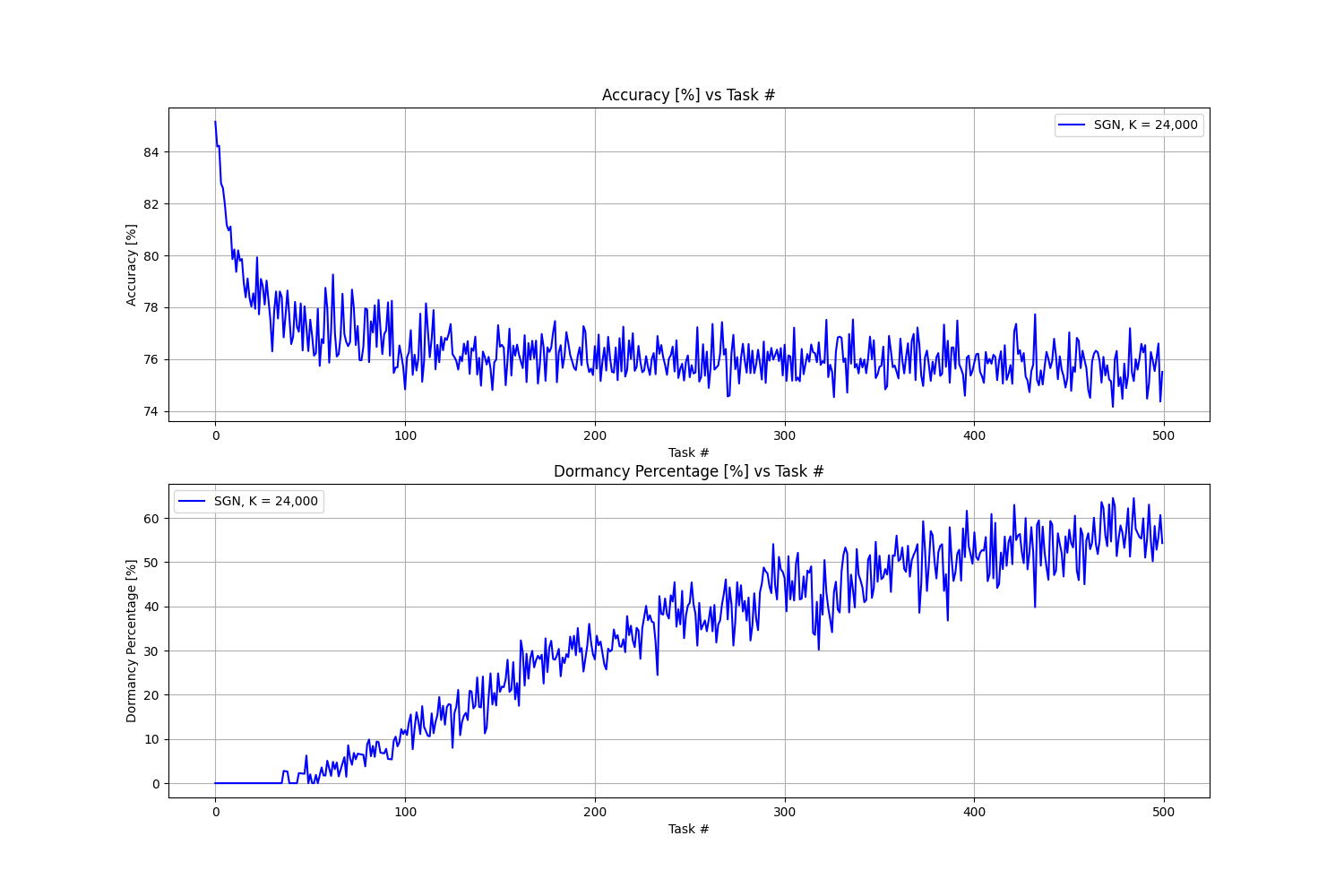}
    \caption{Accuracy and dormancy \% plots for SGN with $K = 24,000$ (one new hidden unit per task).}
    \label{fig:SCN_acc_mnist40k}
\end{figure}

\begin{figure}[H]
    \centering
    \vspace{0.0in}    \includegraphics[width=0.95\linewidth,trim=0 15mm 0 15mm,clip]{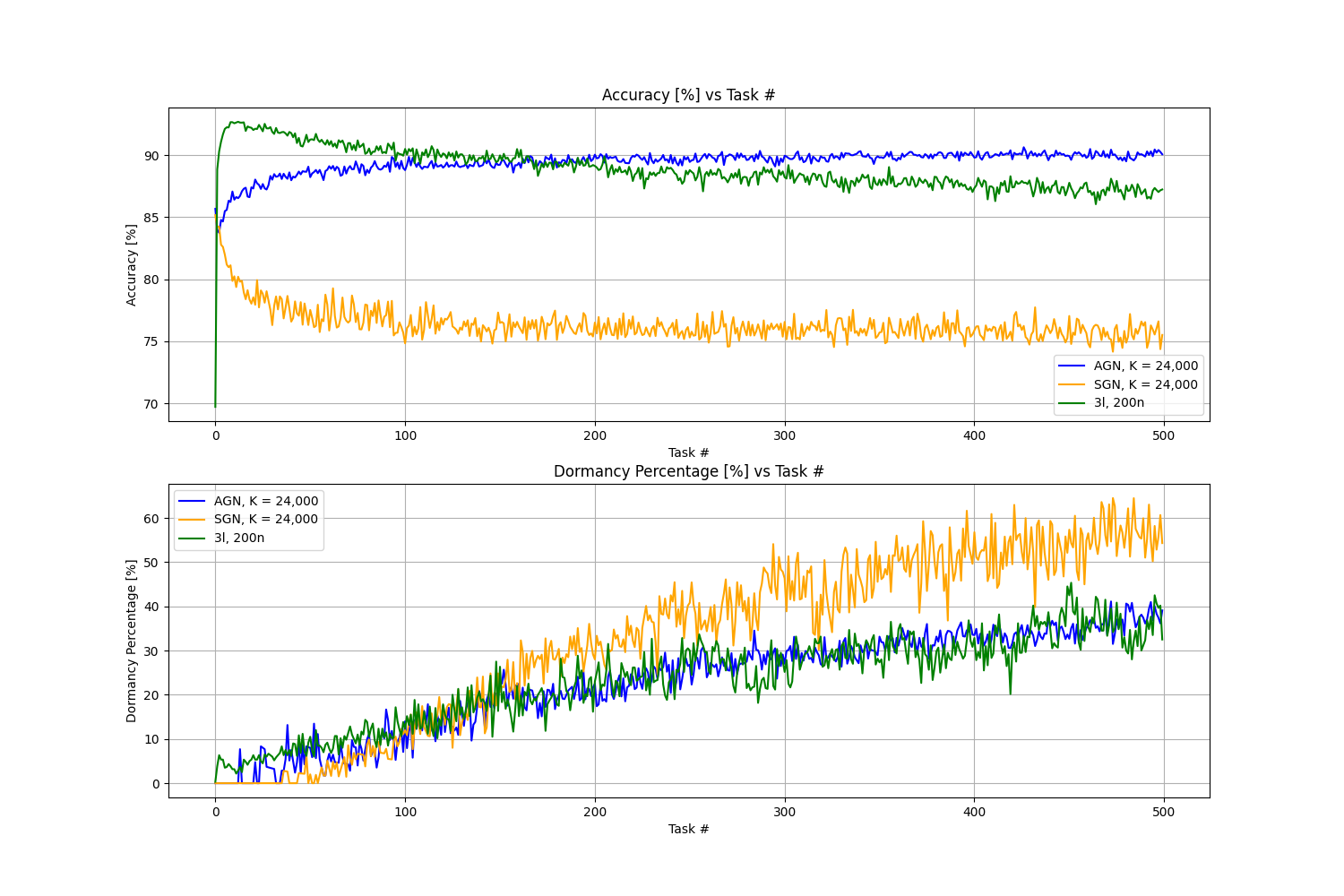}
    \caption{Accuracy and dormancy \% plots for AGN and SGN with $K = 24,000$ (one new hidden unit per task), and F-FCNN with 3 hidden layers and a hidden size of 200.}
    \label{fig:ACN_acc_mnist40k}
\end{figure}

\begin{figure}[H]
    \centering
    \vspace{0.0in}    \includegraphics[width=0.95\linewidth,trim=0 15mm 0 15mm,clip]{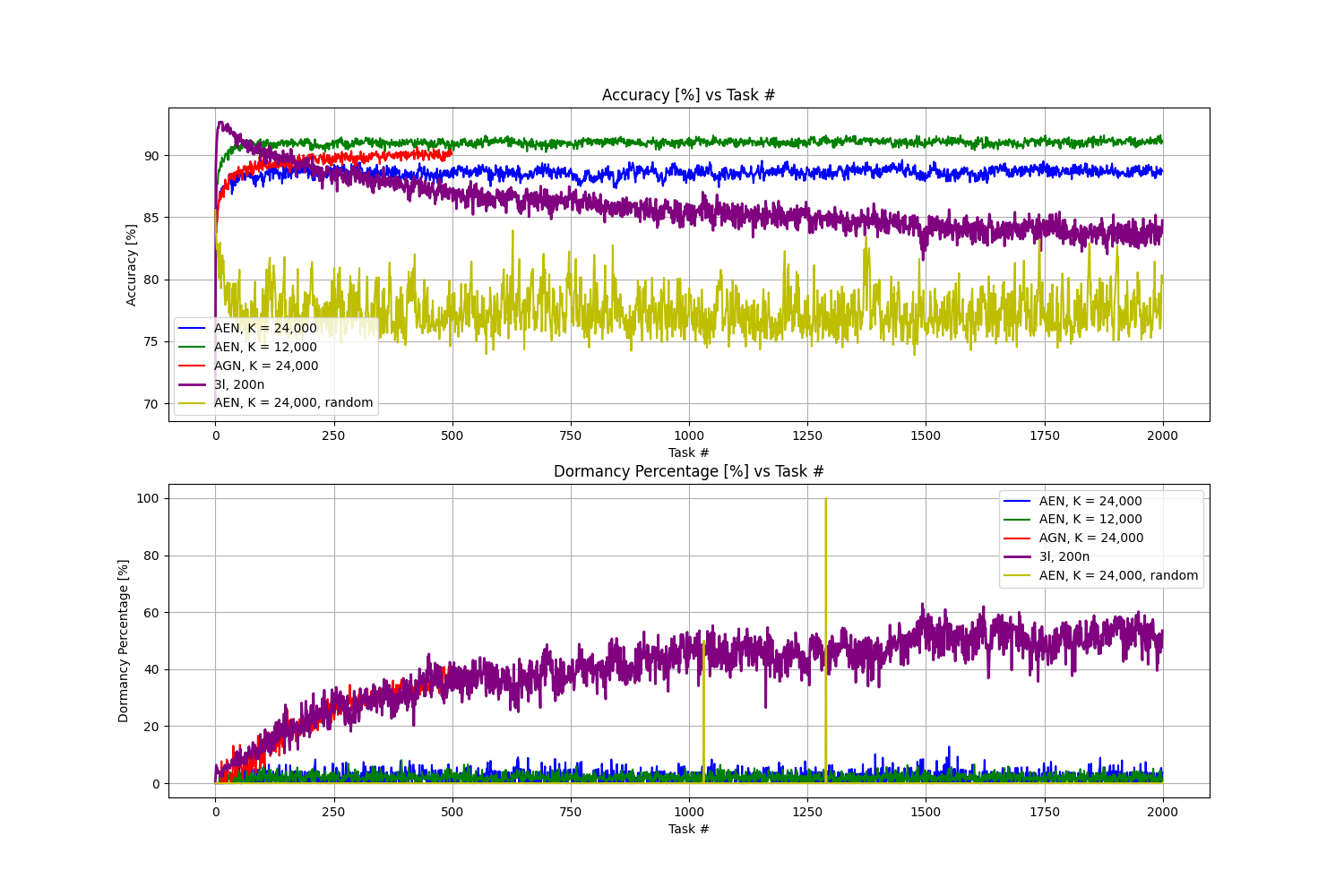}
    \caption{Accuracy and dormancy \% plots for AENs with $K = 24,000$ (one new hidden unit per task) and $K = 12,000$ (three new hidden units per task), along with several baselines.}
    \label{fig:SACN_acc_mnist40k}
\end{figure}

\begin{figure}[H]
    \centering
    \vspace{0.0in}    \includegraphics[width=0.95\linewidth,trim=0 15mm 0 15mm,clip]{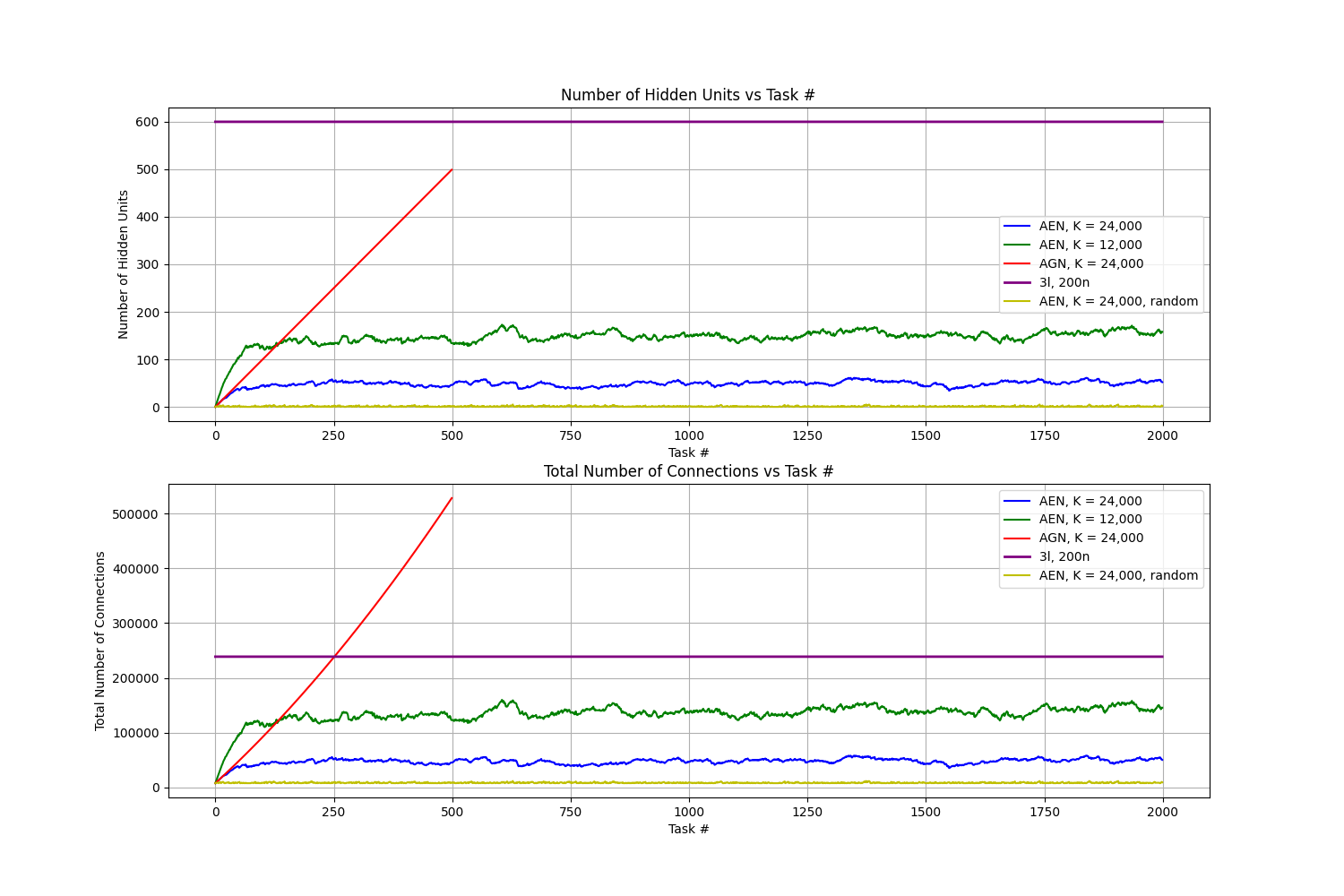}
    \caption{Number of hidden units and number of connections plots for AENs with $K = 24,000$ (one new hidden unit per task) and $K = 12,000$ (three new hidden units per task), along with several baselines.}
    \label{fig:SACN_size_mnist40k}
\end{figure}

\begin{figure}[H]
    \centering
    \vspace{0.0in}    \includegraphics[width=0.73\linewidth,trim=0 35mm 0 35mm,clip]{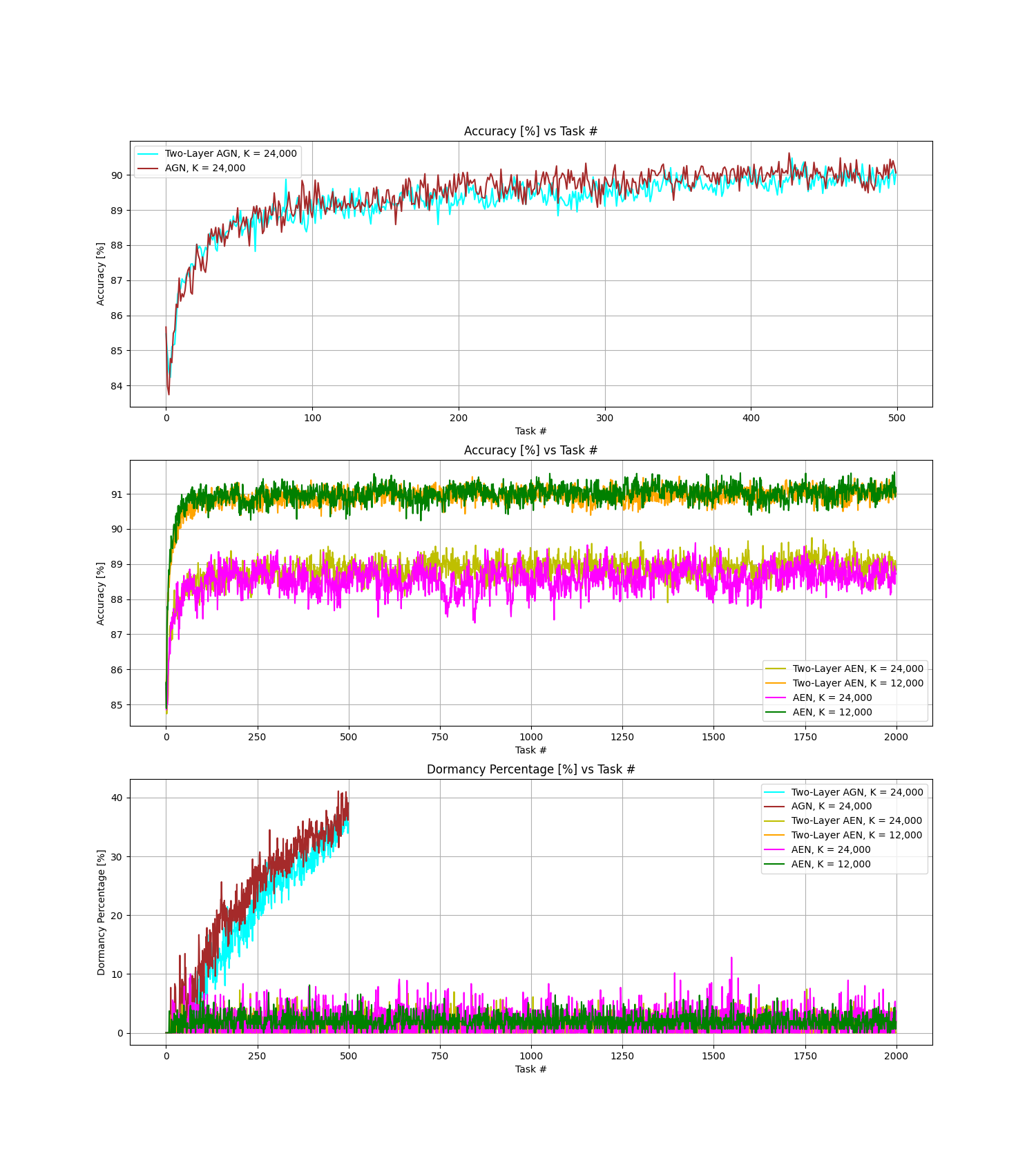}
    \caption{Accuracy and dormancy \% plots for two-layer AGN with $K = 24,000$ (one new hidden unit per task), two-layer AENs with $K = 24,000$ and $K = 12,000$ (three new hidden units per task), along with their original counterparts.}
    \label{fig:two-layer_acc_mnist40k}
\end{figure}

\begin{figure}[H]
    \centering
    \vspace{0.0in}    \includegraphics[width=0.73\linewidth,trim=0 15mm 0 15mm,clip]{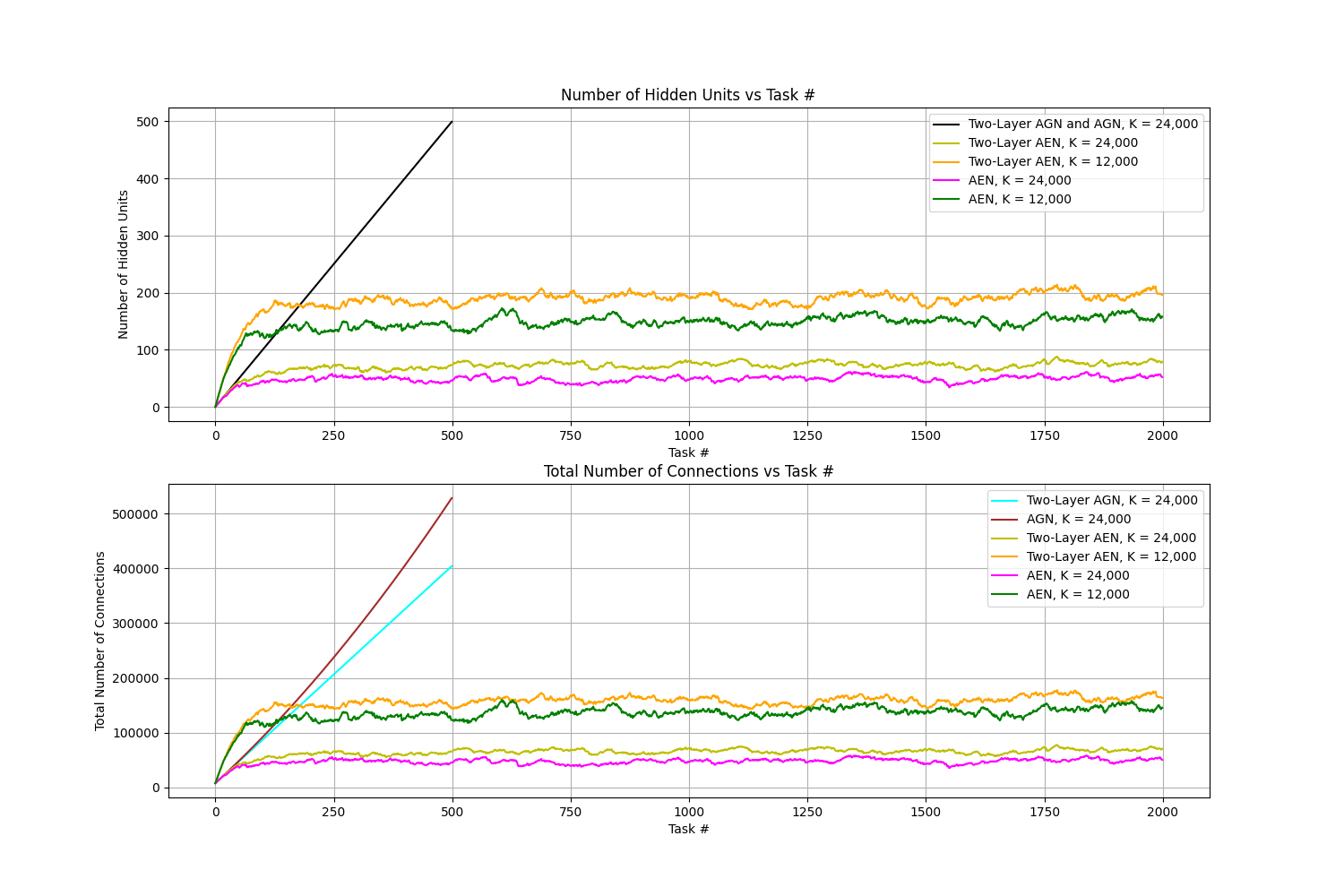}
    \caption{Number of hidden units and number of connections plots for two-layer AGN with $K = 24,000$ (one new hidden unit per task), two-layer AENs with $K = 24,000$ and $K = 12,000$ (three new hidden units per task), along with their original counterparts.}
    \label{fig:two-layer_size_mnist40k}
\end{figure}

\clearpage
\section{Experimental Results for Online Permuted FMNIST, $N = 40,000$}
\label{sec:appendixC}

\begin{figure}[H]
    \centering
    \vspace{0.0in}    \includegraphics[width=0.95\linewidth,trim=0 15mm 0 15mm,clip]{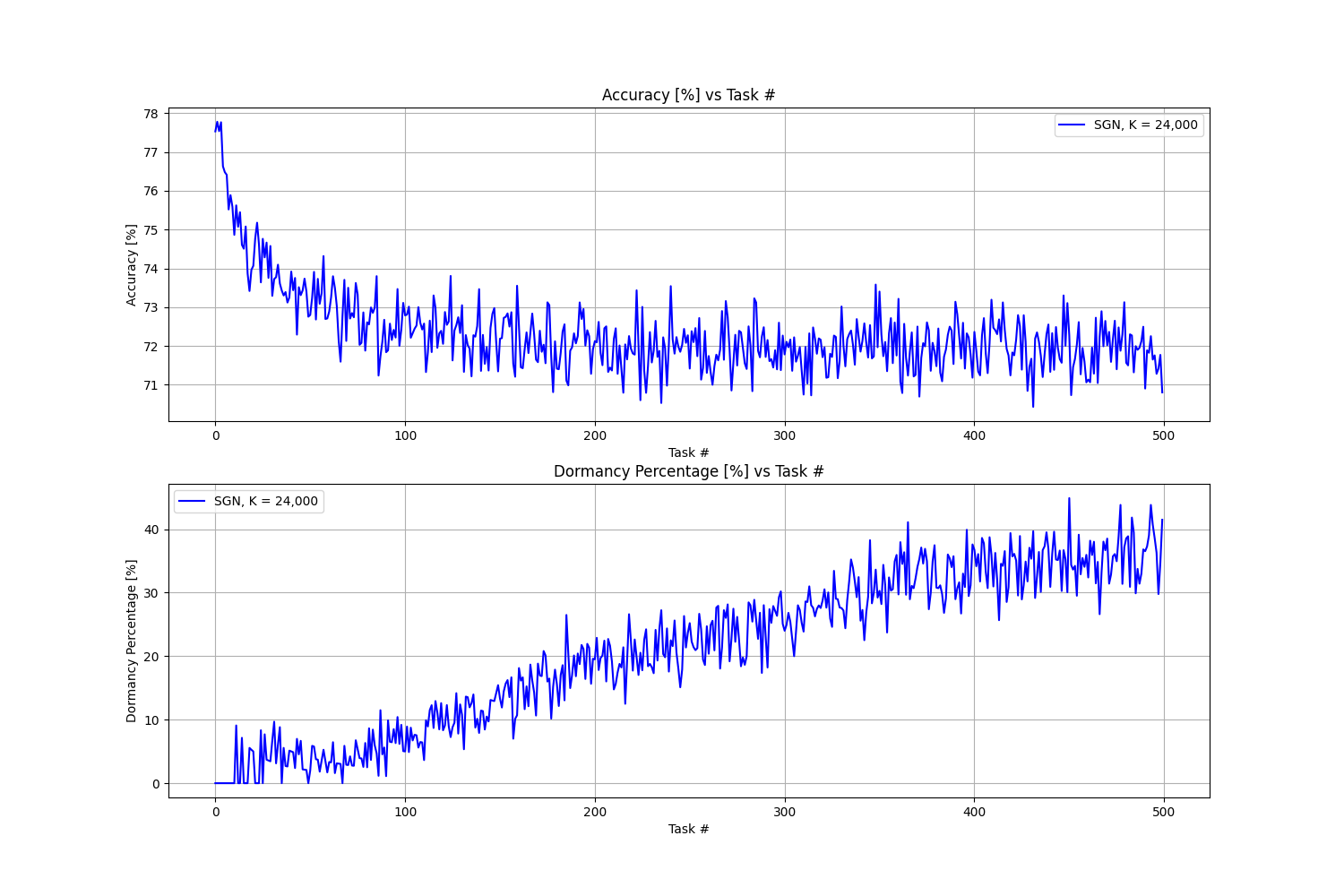}
    \caption{Accuracy and dormancy \% plots for SGN with $K = 24,000$ (one new hidden unit per task).}
    \label{fig:SCN_acc_fmnist40k}
\end{figure}

\begin{figure}[H]
    \centering
    \vspace{0.0in}    \includegraphics[width=0.95\linewidth,trim=0 15mm 0 15mm,clip]{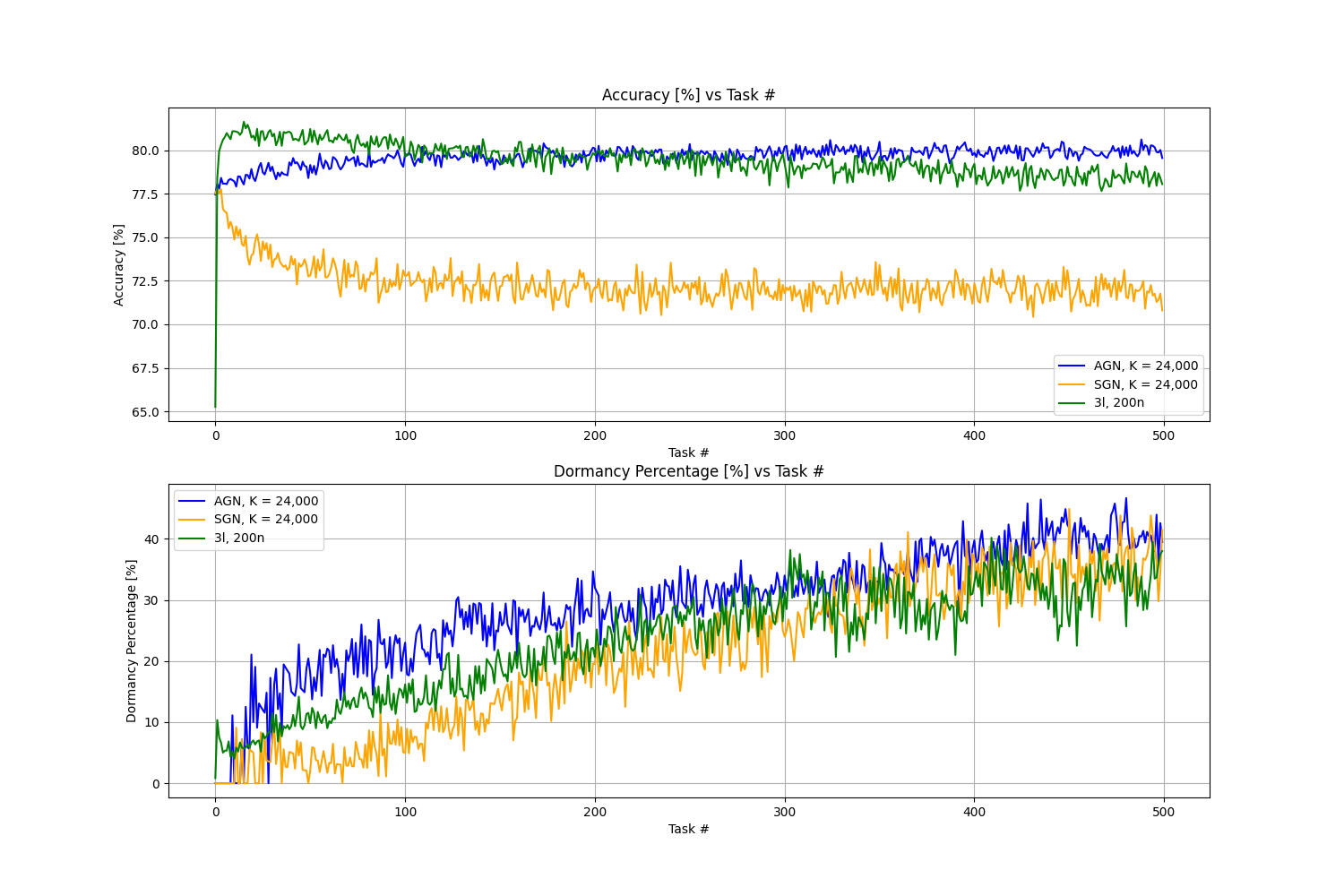}
    \caption{Accuracy and dormancy \% plots for AGN and SGN with $K = 24,000$ (one new hidden unit per task), and F-FCNN with 3 hidden layers and a hidden size of 200.}
    \label{fig:ACN_acc_fmnist40k}
\end{figure}

\begin{figure}[H]
    \centering
    \vspace{0.0in}    \includegraphics[width=0.95\linewidth,trim=0 15mm 0 15mm,clip]{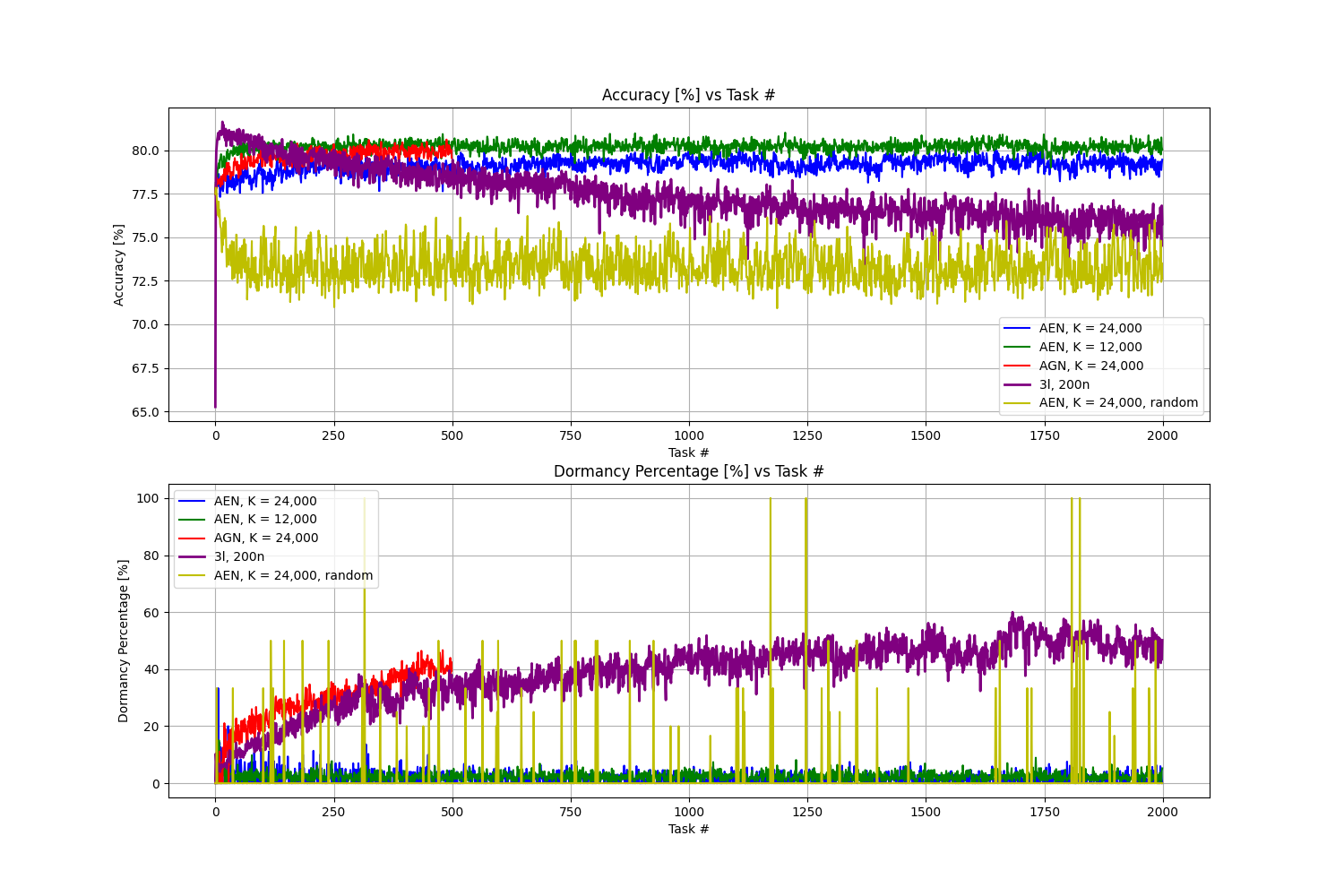}
    \caption{Accuracy and dormancy \% plots for AENs with $K = 24,000$ (one new hidden unit per task) and $K = 12,000$ (three new hidden units per task), along with several baselines.}
    \label{fig:SACN_acc_fmnist40k}
\end{figure}

\begin{figure}[H]
    \centering
    \vspace{0.0in}    \includegraphics[width=0.95\linewidth,trim=0 15mm 0 15mm,clip]{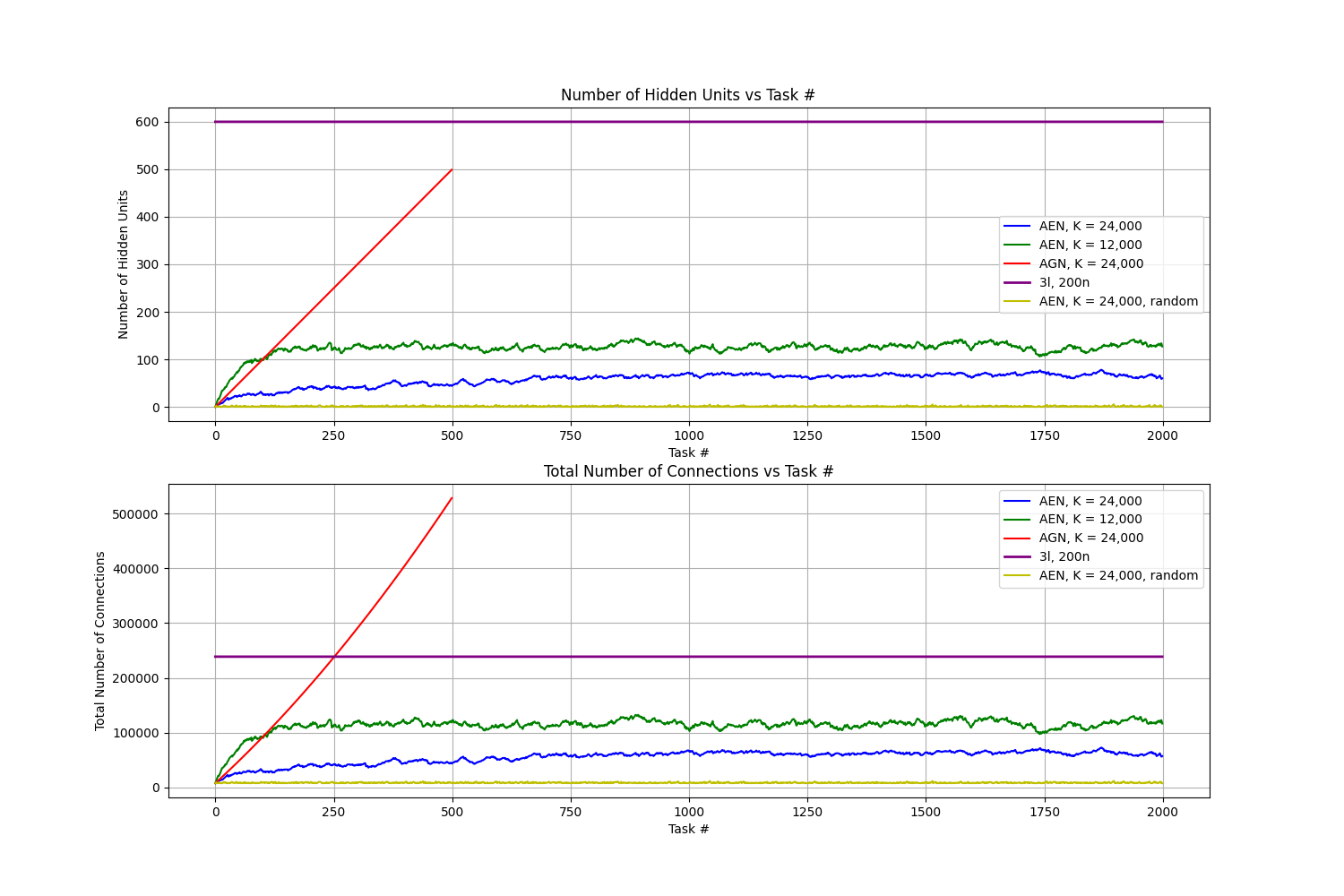}
    \caption{Number of hidden units and number of connections plots for AENs with $K = 24,000$ (one new hidden unit per task) and $K = 12,000$ (three new hidden units per task), along with several baselines.}
    \label{fig:SACN_size_fmnist40k}
\end{figure}

\begin{figure}[H]
    \centering
    \vspace{0.0in}    \includegraphics[width=0.73\linewidth,trim=0 35mm 0 35mm,clip]{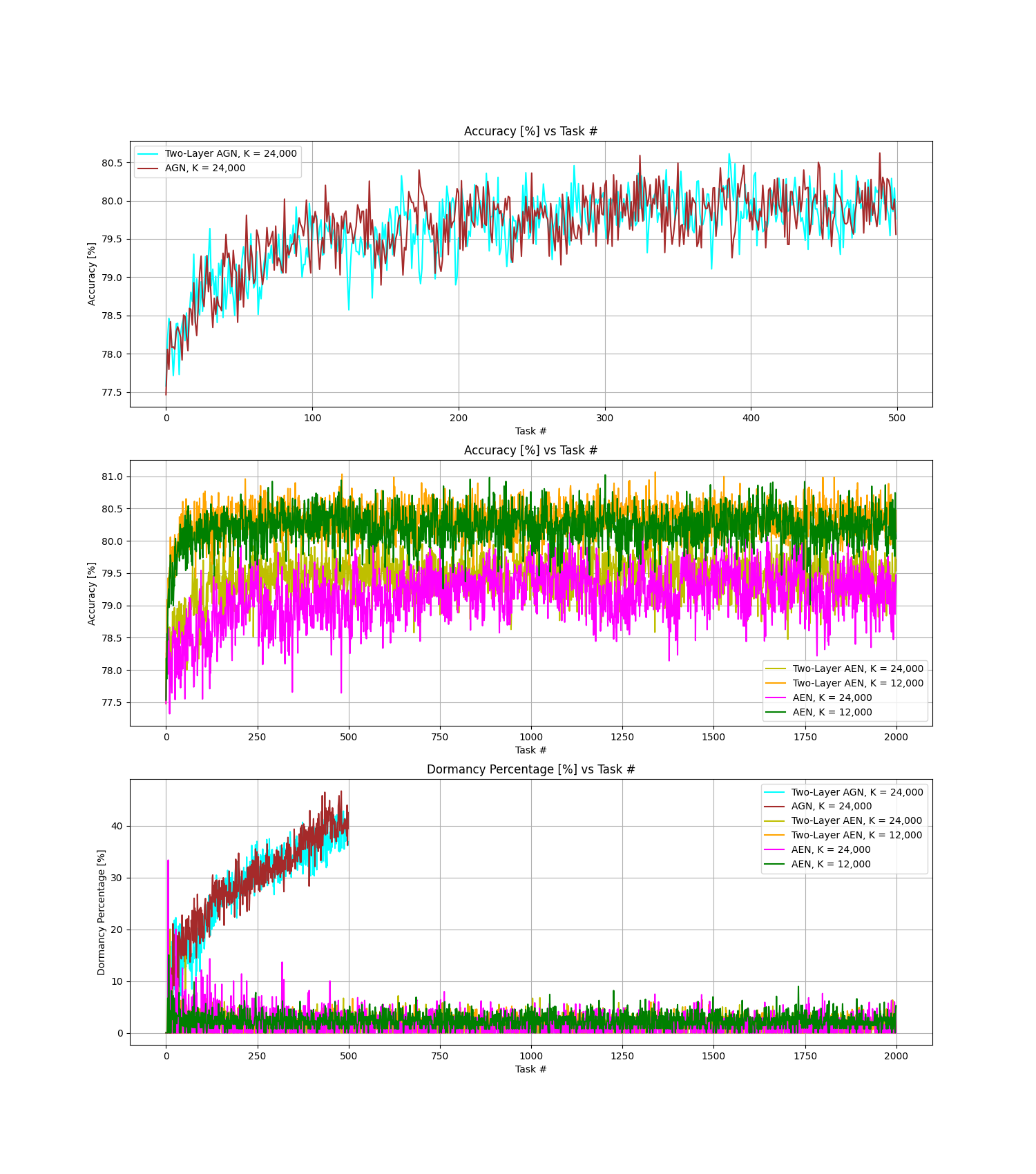}
    \caption{Accuracy and dormancy \% plots for two-layer AGN with $K = 24,000$ (one new hidden unit per task), two-layer AENs with $K = 24,000$ and $K = 12,000$ (three new hidden units per task), along with their original counterparts.}
    \label{fig:two-layer_acc_fmnist40k}
\end{figure}

\begin{figure}[H]
    \centering
    \vspace{0.0in}    \includegraphics[width=0.73\linewidth,trim=0 15mm 0 15mm,clip]{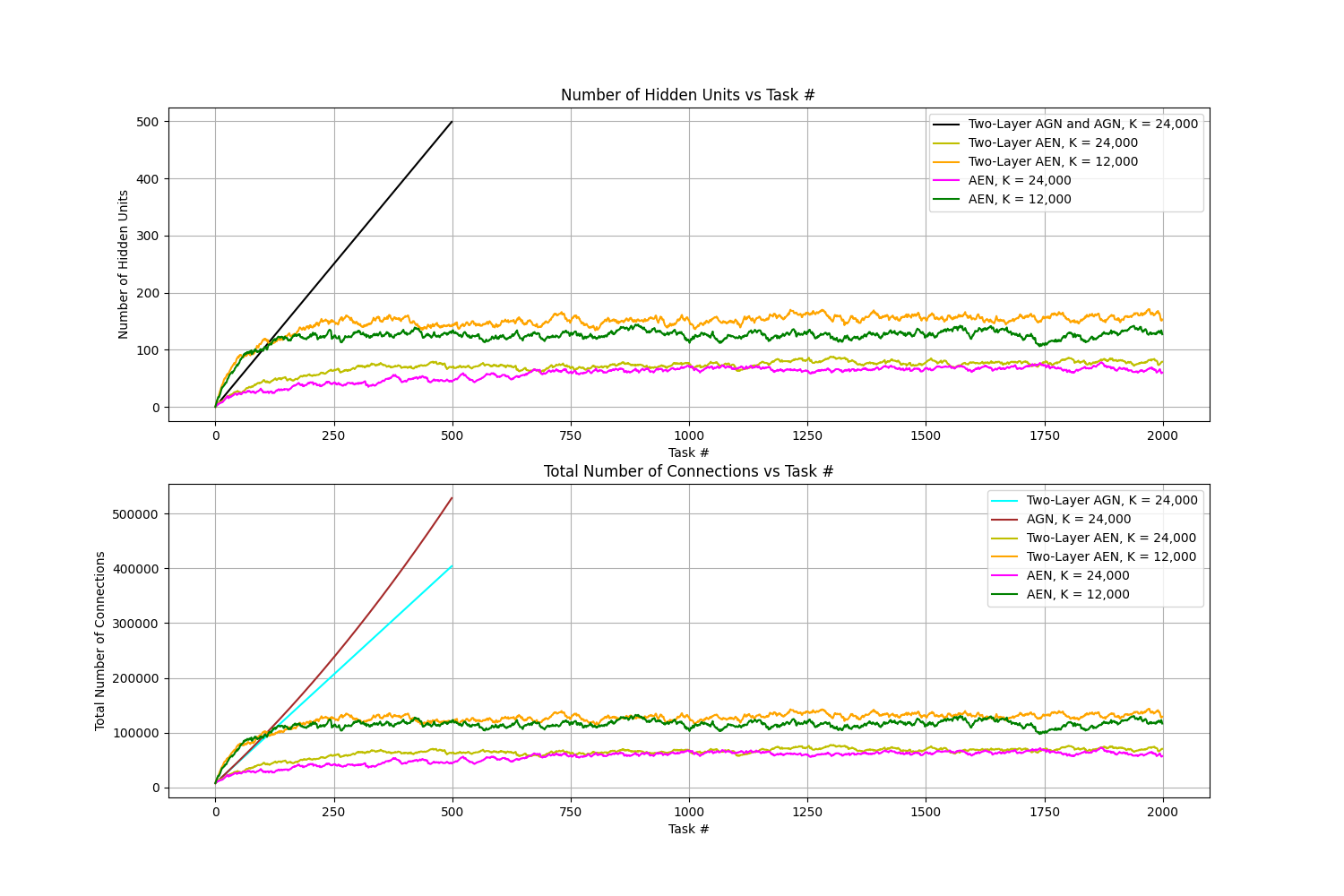}
    \caption{Number of hidden units and number of connections plots for two-layer AGN with $K = 24,000$ (one new hidden unit per task), two-layer AENs with $K = 24,000$ and $K = 12,000$ (three new hidden units per task), along with their original counterparts.}
    \label{fig:two-layer_size_fmnist40k}
\end{figure}

\end{document}

%% file: math_commands.tex
\usepackage{amsmath,amsfonts,bm}

\def\eqref#1{equation~\ref{#1}}

\def\1{\bm{1}}

\DeclareMathAlphabet{\mathsfit}{\encodingdefault}{\sfdefault}{m}{sl}
\SetMathAlphabet{\mathsfit}{bold}{\encodingdefault}{\sfdefault}{bx}{n}

